\documentclass[a4paper]{article}
\usepackage[utf8]{inputenc}
\usepackage{graphicx} 
\usepackage{authblk} 
\usepackage{amsmath} 
\usepackage{hyperref} 
\usepackage{graphicx, float}
\usepackage{lineno}
\usepackage[a4paper,left=3cm,right=2cm,top=2.5cm,bottom=2.5cm]{geometry}
\newcounter{suppfigure}

\usepackage{xcolor}

\newcommand{\suppautoref}[1]{Supplementary~\autoref{#1}}
\usepackage{textcomp}
\usepackage{newunicodechar}
\newunicodechar{′}{\ensuremath{'} }

\renewcommand{\figurename}{Fig.} 

\usepackage[isbn=false,url=false,eprint=false,sorting=none,citestyle=numeric-comp]{biblatex}
\usepackage{subcaption}

\usepackage{multirow}

\title{Unsupervised local learning based on voltage-dependent synaptic plasticity for resistive and ferroelectric synapses}

\author[1,2,3]{Nikhil Garg}
\author[1]{Ismael Balafrej}
\author[1,2,4,5]{Joao Henrique Quintino Palhares}
\author[6]{Laura Bégon-Lours}
\author[1,2]{Davide Florini}
\author[6]{Donato Francesco Falcone}
\author[6]{Tommaso Stecconi}
\author[6]{Valeria Bragaglia}
\author[6]{Bert Jan Offrein}
\author[7]{Jean-Michel Portal}
\author[8]{Damien Querlioz}
\author[1,2]{Yann Beilliard}
\author[1,2]{Dominique Drouin}
\author[1,2,3]{Fabien Alibart\setcounter{footnote}{0}\footnote{Corresponding author. \newline 
Fabien.Alibart@Usherbrooke.ca}}

\affil[1]{Institut Interdisciplinaire d′Innovation Technologique (3IT), Université de Sherbrooke, 3000 Boulevard de l'université, Sherbrooke, J1K OA5 Québec, Canada}
\affil[2]{Laboratoire Nanotechnologies Nanosystèmes (LN2)-IRL3463, CNRS, Université de Sherbrooke, INSA Lyon, Ecole Centrale de Lyon, Université Grenoble Alpes, Sherbrooke, J1K 0A5 Québec, Canada}
\affil[3]{Institute of Electronics, Microelectronics and Nanotechnology (IEMN), Université de Lille, 59650 Villeneuve d’Ascq, France}
\affil[4]{STMicroelectronics, 850 rue Jean Monnet, 38920 Crolles, France}
\affil[5]{Univ. Grenoble Alpes, CEA, CNRS, Grenoble INP, SPINTEC, 38000 Grenoble, France}
\affil[6]{IBM Research GmbH - Zurich Research Laboratory, CH-8803 Ruschlikon, Switzerland}
\affil[7]{Aix Marseille Univ, Université de Toulon, CNRS, IM2NP, Marseille, France}
\affil[8]{Université Paris-Saclay, CNRS, Centre de Nanosciences et de Nanotechnologies, Palaiseau, France}

\date{}

\begin{document}
\maketitle

{\noindent\textbf{Abstract}
The deployment of AI on edge computing devices faces significant challenges related to energy consumption and functionality. These devices could greatly benefit from brain-inspired learning mechanisms, allowing for real-time adaptation while using low-power. In-memory computing with nanoscale resistive memories may play a crucial role in enabling the execution of AI workloads on these edge devices. In this study, we introduce voltage-dependent synaptic plasticity (VDSP) as an efficient approach for unsupervised and local learning in memristive synapses based on Hebbian principles. This method enables online learning without requiring complex pulse-shaping circuits typically necessary for spike-timing-dependent plasticity (STDP). We show how VDSP can be advantageously adapted to  three types of memristive devices (TiO$_2$, HfO$_2$-based metal-oxide filamentary synapses, and HfZrO$_4$-based ferroelectric tunnel junctions (FTJ)) with disctinctive switching characteristics. System-level simulations of spiking neural networks incorporating these devices were conducted to validate unsupervised learning on MNIST-based pattern recognition tasks, achieving state-of-the-art performance. The results demonstrated over 83\% accuracy across all devices using 200 neurons. Additionally, we assessed the impact of device variability, such as switching thresholds and ratios between high and low resistance state levels, and proposed mitigation strategies to enhance robustness.
\\
}

\section{Introduction}\label{sec1}

Deploying artificial intelligence (AI) applications on edge computing devices is raising the challenge of implementing intelligent algorithms with sever constraints on energy consumption, challenge that cannot be fulfilled by conventional technologies such as modern GPU. One strategy toward this goal is relying on the neuromorphic engineering and computing framework, which proposes the physical implementation of algorithms with specialized hardware designed to emulate the human brain's structure and function. Among the different propositions of neuromorphic devices and circuits,  memristors \cite{chua1971memristor, strukov2008missing} —resistive devices with programmable conductance— have been considered as emerging memory devices used to create electronic synapses~\cite{jo2010nanoscale} in AI hardware systems. In neuromorphic architectures, memristors enable the realization of the Vector Matrix Multiplication function physically through Ohm's and Kirchoff`s laws, which reduces significantly energy consumption and latency of the intensive VMM operation. In addition, memristor have generated a large interest to implement physically the different learning algorithms required during training of neuromorphic systems. Learning algorithms used for conventional artificial neural networks (ANNs) mostly rely on backpropagation and have been widely considered in the context of in-memory computing with memristor with severe constraints on memristors’ accuracy (i.e. number of states available during programming), linearity (i.e. linear change of conductance on the entire resistive range) and variability \cite{hasan2017chip, alibart2013pattern, snider2007self, burr2015experimental}. In neuromorphic approaches, and in particular in spiking neural networks (SNNs) \cite{maass1997networks}, learning algorithms rely deeply on bio-inspiration and memristors could offer the additional advantage of local learning by implementing Hebbian principles. For instance, Spike timing-dependent plasticity (STDP) \cite{bi1998synaptic} have been found to be responsible for plasticity in biological synapses and worth adapting to hardware systems \cite{arthur2005learning}. In such learning algorithms, change of states of the synaptic conductance depends only on the pre and post neuron activity and doesn’t require the propagation of global error signals across the entire network, thus limiting data movement and the associated energy consumption. Various STDP implementations have considered memristor to implement online learning in SNN but (i) translation of pre- and post-neuron activity into actual signals promoting the change of resistance (i.e. learning) is impacting the overall network performances and (ii) the impact of memristive devices non-idealities is not straightforward to extract and still largely unexplored with respect to the large variety of memristive devices that could be considered based on physical mechanisms such as heating~\cite{fursina2009origin}, oxidation~\cite{waser2009redox, nian2007evidence}, phase change~\cite{burr2010phase, wuttig2007phase}, and ferroelectric domain~\cite{chanthbouala2012ferroelectric} switching. 

In this paper, we propose (i) to adapt a recent extension of STDP to memristive devices and to evaluate its performances on a relevant classification task. We consider for this study Voltage Dependent Synaptic Plasticity (VDSP) \cite{garg2022voltage, goupy2023unsupervised}, which uses spike timing and neuron membrane potential as a representation of pre and post neuron activities. Such local learning algorithm has been proposed recently as an interesting solution to solve the strong dependency of STDP to the range of frequency in SNNs and to ease the physical implementation by reducing the number of local parameters to be stored. (ii) We further analyse the impact of memristive devices properties on the performances and resilience of VDSP. We consider three distinctive technologies with different switching dynamics associated to different physical mechanisms originating the change of resistance. TiO$_2$ and Conductive-Metal-Oxide (CMO)-HfO$_2$ are representative of valence change memory devices where oxygen vacancies are responsible for resistive switching and HZO represents the more recent class of ferroelectric tunnel junctions where ferroelectric domains are defining the resistive states. These choice of the device stack was motivated by their CMOS-compatible fabrication process \cite{el2022fully, begon2022scaled, falcone2023physical} for back end of line (BEOL) integration. All three device can present analog change of resistance that could be advantageously used for online learning implementation, but with different relationship in between driving signals and devices’ response (i.e. voltage driven resistive change in our case). We show in this paper how VDSP parameters can be adjusted to each technology by combining electrical characterization of the different technologies and electrical modeling. While individual technologies evaluation is often reported and can limit the generalization of the impact of device non-idealities on learning, we show here for three technologies how important switching characteristics such as variation in switching threshold and range of resistance changes can affect the performances of VDSP.

Several works have proposed implementation of STDP with memristive devices~\cite{feldman2012spike}. In these approaches, temporal correlation in between pre- and post- neuron events can be advantageously translated into voltage with pulse overlapping technique~\cite{serrano2013stdp}. This technique enables the implementation of online learning in various memristive device technologies and has been proven efficient for unsupervised pattern learning and circumventing known issues of variability and stochasticity in memristive devices \cite{querlioz2013immunity, yu2011stochastic}.  Nevertheless, such pulse-overlapping method for temporal correlation detection faces challenges since the duration of the programming pulse is directly related to the time window for detecting the correlated spikes.  When implemented into circuits, long pulses come with trade-offs such as reduced analog control, increased energy consumption for programming, and lower throughput. Additionally, if complex pulses (i.e. with exponential decaying function) can translate time distance in between the pre and post events into a large range of voltages, such pulses engineering is associated to complex voltage sources design that need to be adapted to each different learning algorithm and memristive technology, thus preventing a general circuit design approach.  

From a system level perspective, conventional STDP requires to store locally at the pre and post neuron level the timing of the last emitted spike (note that pulse overlapping techniques are circumventing this issue by storing the pulse timing into the synaptic programming voltage). Strategies to reduce the local memory requirement have been proposed with neuron-state-dependent synaptic plasticity for CMOS synapses \cite{mitra2008real}. These learning rules, known as spike-driven synaptic plasticity (SDSP) \cite{frenkel2017fully}, or single-spike STDP, use the membrane potential of the post-synaptic neuron and an extra calcium concentration variable associated with the rate of spike of the post-neuron to modulate the synaptic weight during a pre-synaptic spike, thus reproducing Hebbian learning concept. VDSP \cite{garg2022voltage, goupy2023unsupervised} employ a similar strategy by extracting the probability of pre-neuron firing from the neuron membrane potential. This approach further reduces the memory requirement by removing the calcium concentration memory block, while still maintaining a high-performing network, as demonstrated through recognition rates of handwritten digits obtained through unsupervised learning.

In the following, we first introduce the theoretical framework for implementing VDSP in analog memristive synapses. We propose a device characterization approach that models the programming behavior as a function of the applied voltage and the current device state. This behavior is used to fit a standard memristor model, allowing us to quantify key switching properties such as the switching threshold and nonlinear characteristics. Next, we discuss the mapping from device characteristics to simulation and outline the SNN simulation setup. Performance is incrementally evaluated based on key factors, including the number of output neurons, training samples, and learning epochs. Finally, we examine the impact of device variation and present hardware strategies to mitigate performance degradation, showing improvements by a significant margin. 

\section{Results}

\subsection{Voltage-dependent switching of memristors}

\begin{figure}[H]
     \centering
         \includegraphics[clip,width=18cm,height=10cm,keepaspectratio, width=1\textwidth]{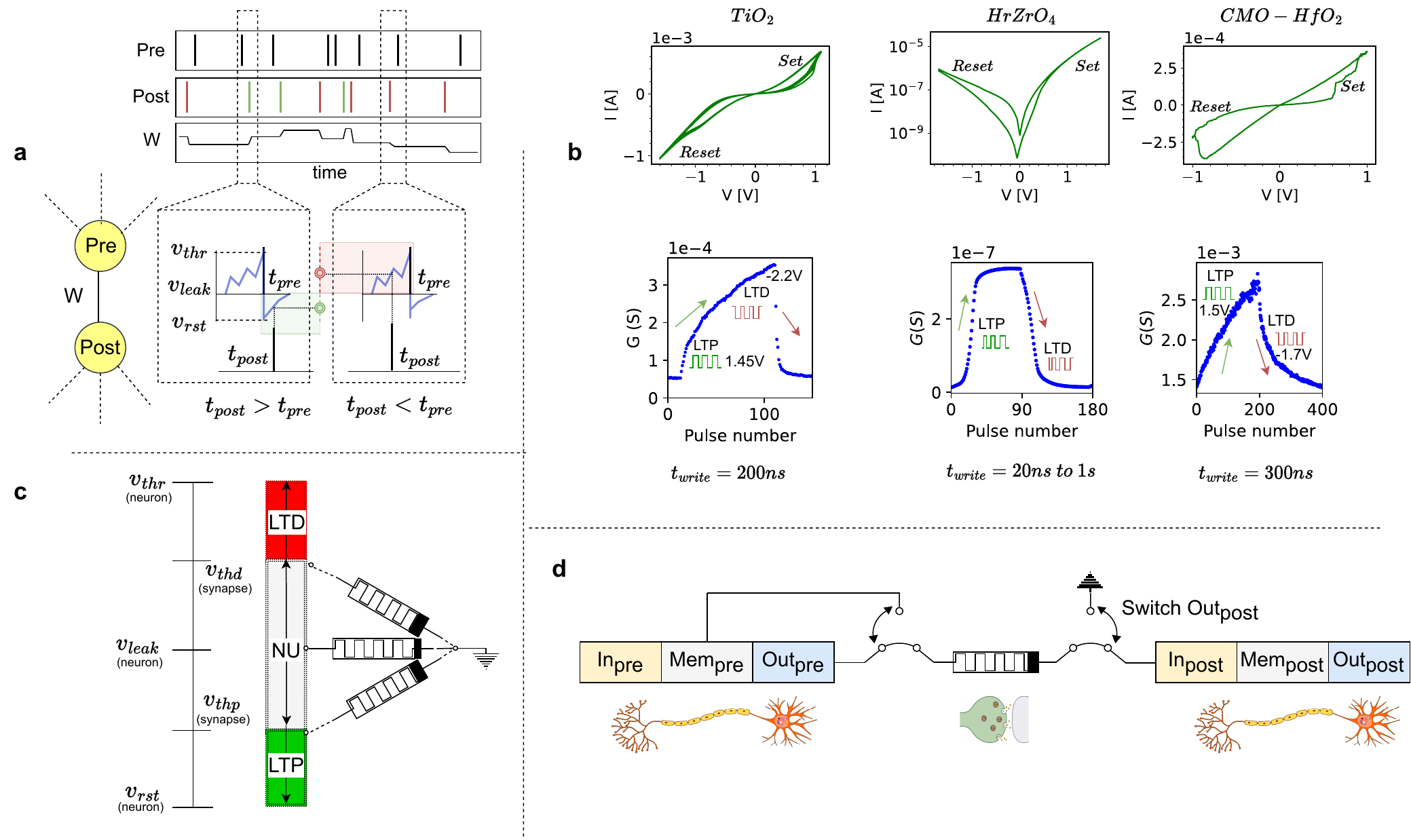}
    \caption{Schematic representation of the VDSP learning rule implemented in a memristive synapse between a pre- and postsynaptic spiking neuron. 
    {\footnotesize \textbf{a} Theoretical framework of VDSP in determining the causality of pre- and post-neuron spike events. The synapse is potentiated when the post-neuron spikes after the pre-neuron, and is depressed when the pre-neuron spikes following the post-neuron spike event. \textbf{b} Cumulative long term potentiation/depression (LTP/LTD) plot obtained with the response by pulses of positive polarity to show LTP and fixed magnitude followed by ones with negative polarity to induce LTD. DC sweeps (I-V) characteristics loops for all three devices is shown below. \textbf{c} Regions for long-term potentiation (LTP), long-term depression (LTD), and no update (NU) are based on voltage-regulated, threshold-dependent memristor switching. Schematic representation of VDSP for memristor programming. \textbf{d} Circuit-level depiction of VDSP on a single memristive synapse. The output from the post-synaptic neuron controls the switch between the inference and weight update phases. In, Mem, and Out represent the input terminal, membrane potential, and output terminal respectively for pre- and post-synaptic neuron.
}}
    \label{fig:article2-fig1}
\end{figure}

Synaptic memory, based on Hebbian learning principles, captures the history of causal and anti-causal spike pairs between the pre-neuron and post-neuron. Using voltage-dependent synaptic plasticity (VDSP), the recent activity of the pre-neuron can be inferred from the neuron's membrane potential. A low membrane potential corresponds to a recent firing event, while a high membrane potential signals an imminent spike, as illustrated in \autoref{fig:article2-fig1}a. This learning mechanism is translated into the programming strategy for memristors, where synaptic conductance modulation is stored as a function of the history of applied voltages during successive spiking events.

In memristive devices, switching is primarily controlled by the programming voltage amplitude and duration. The distinctive current-voltage (I-V) characteristics (pinched hysteresis loop) of the three devices are shown in \autoref{fig:article2-fig1}b (top panels). These measurements are representative single device current-voltage sweeps that highlight the distinctive pinched hysteresis loops of the three devices. In bipolar memristive devices, the hysteresis loop evidences the High Resistance State (HRS) and Low Resistance State (LRS) achieved after applying a voltage of opposite polarity. HRS and LRS define the switching range of the device in its binary regime. All three devices show very distinctive I-V hysteresis signatures that are linked to the physical mechanisms involved during switching. In oxide-based memristive devices, such behavior is intimately linked to the balance between drift and diffusion \cite{sassine2016interfacial} and results in different voltage-resistance dependencies. In ferroelectric devices, switching dependency is associated with nucleation mechanisms in ferroelectric domains that govern the resistance states of the tunnel junction \cite{chanthbouala2012ferroelectric, chanthbouala2012solid}. 

\autoref{fig:article2-fig1}b (bottom) illustrates how switching can be controlled in an analog way when the programming voltage is applied as a sequence of pulses. The gradual change of conductance during the increase of resistance (LTD) and decrease of resistance (LTP) can be advantageously used to implement online learning. In this example, LTP and LTD are obtained with constant amplitude pulses, and the transitions evidence the cumulative effects that can be obtained by adjusting only the pulse duration. This scenario (i.e., gradual switching through cumulative effects of identical pulses) is the most straightforward way to implement learning in ANNs and SNNs, as each learning event can be associated with the application of a single pulse without adapting the pulse shape, and actual learning results from the repetition of the same learning signal. More complex situations occur when both pulse amplitude and pulse duration can be modulated during the learning signal, with the benefit of a larger dynamic range of programming and a higher number of states available \cite{alibart2012high}.

From \autoref{fig:article2-fig1}b (bottom), all three technologies present different signatures in their analog regimes that can be further analyzed in terms of the number of states available, linearity of the transition, and min/max resistance states. For instance, implementing online learning in ANNs would favor the highest number of states between the HRS and LRS, the most linear transition, and the least variability between the HRS and LRS of different devices to map the backpropagated error signal with the highest accuracy. In the context of SNNs, the impact of these parameters is less clear and needs to be evaluated for each learning scenario. For example, conventional STDP could translate the magnitude of the LTP/LTD into different pulse durations \cite{ilyas2020analog} or a combination of pulse duration and pulse amplitude \cite{campbell2016pulse}. 

VDSP relies on the internal membrane voltage parameters as a probability of a spike being correlated or anti-correlated. The magnitude of the pre-neuron membrane voltage potential is directly associated with the magnitude of the learning signal when a post-synaptic spike is emitted. \autoref{fig:article2-fig1}c illustrates how the membrane voltage potential of the neuron can be translated into a programming voltage of the memristor. The neuron's membrane potential, which lies between the threshold (V\textsubscript{thr}) and reset potential (V\textsubscript{rst}), can be mapped to the min/max voltage applied to the memristor, respectively. Such mapping results in a non-switching region for small voltages (when events are not strongly correlated) and LTD/LTP events when the voltage is above/below the switching threshold of the memristor.

This solution greatly simplifies the programming scenario and offers two main advantages for online learning implementation: (i) The membrane potential of the neuron can be converted into sub-microsecond pulses for memristor programming while capturing spike coincidences over millisecond-long windows based on the membrane potential time constant. Using short pulses allows for more precise analog control of memristor conductance switching and reduces energy consumption. (ii) Since the neuron's membrane potential provides a range of programming voltages, it can fully utilize the entire conductance range of the devices. This contrasts with fixed voltage pulse programming, where the ON/OFF ratio is often compromised to achieve gradual conductance modulation.

\subsection{Electrical characterization}

\begin{figure}[H]
     \centering
         \includegraphics[clip,width=18cm,height=10cm,keepaspectratio, width=1\textwidth]{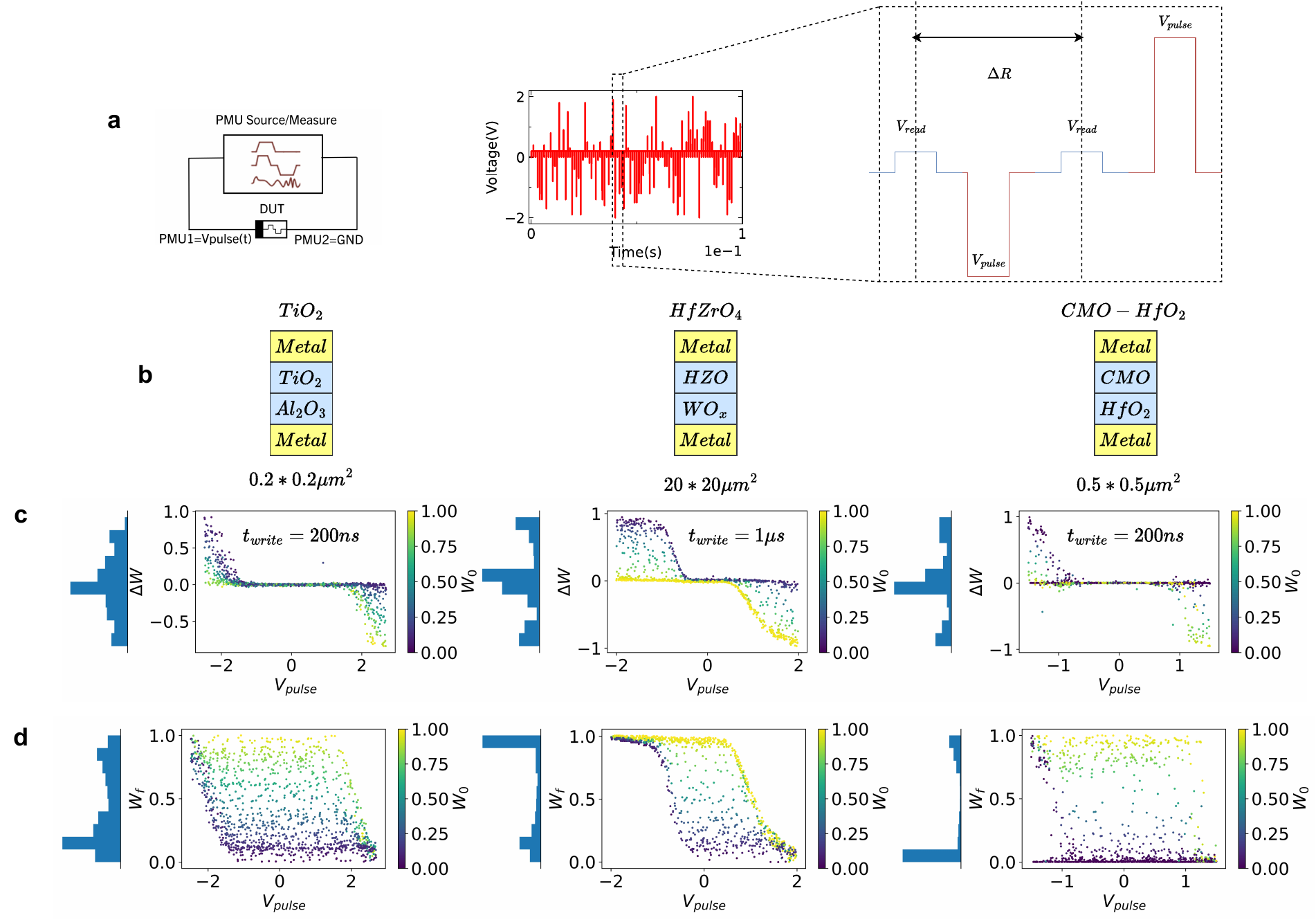}
   \caption{Device characterization protocol, stack, and switching behavior.
    {\footnotesize
    \textbf{a} The characterization protocol involves applying write pulses of random amplitude with a constant pulse width, followed by read pulses. 
    \textbf{b} The device stack for the three characterized devices includes metal top and bottom electrodes and switching oxide layers in between. 
    \textbf{c} Weight change ($\Delta W$) in relation to the applied programming pulse ($V_{pulse}$) and the initial weight ($W_0$) for the three device stacks: $TiO_2$, HZO, and CMO-HfO$_2$, from left to right. 
    \textbf{d} Final weight ($W_f$) as a function of the applied voltage ($V_{pulse}$) and initial weight across the investigated device stacks. Histograms in (c) and (d) represent the change of weight and final weight, respectively, for the entire pulse protocol.
    }}
    \label{fig:article2-fig2}
\end{figure}

A specialized electrical characterization protocol was developed to characterize and model the voltage-dependent switching behavior. The dynamics of this switching were evaluated by applying short pulses (200ns or 1$\mu$s) at different voltage levels randomly distributed between $V_{min}$ and $V_{max}$ (\autoref{fig:article2-fig2}a). Between each write pulse, the device's resistance was measured with a low-magnitude reading pulse. Using a random sequence helps to explore simultaneously the contribution of voltage amplitude and the impact of the initial state, as initial states are, in principle, randomly generated during the overall sequence. The weight change ($\Delta W$) is plotted as a function of the applied voltage ($V_{pulse}$) and the initial weight ($W_o$) for the three device stacks (\autoref{fig:article2-fig2}c). The weight ($w$) represents the normalized conductance of the device and is calculated as follows:

\begin{equation}
w = \frac{g - g_{\text{min}}}{g_{\text{max}} - g_{\text{min}}}
\label{placeholder}
\end{equation}

Where $g_{min}$ and $g_{max}$ represent the conductance in the HRS and LRS, respectively. The device stacks of the three different devices are depicted in \autoref{fig:article2-fig2}b and the fabrication recipe is detailed later in the Methods section. Histograms in \autoref{fig:article2-fig2}c show the cumulative count of $\Delta W$ over the entire experiment. For the three devices, the highest probability of $\Delta W$ is centered around 0, which can be attributed to the absence of switching when voltages are below the SET and RESET threshold voltages (i.e., dead zone). Here SET refers to the process of switching the memristor to a low-resistance state (w=1), while RESET switches it to a high-resistance state (w=0). TiO$_2$ memories exhibit a more uniform distribution of weight change values compared to HZO and CMO-HfO$_2$, highlighting that this technology presents a more gradual switching for the given pulse amplitude and pulse duration chosen during the protocol (i.e., HZO and CMO-HfO$_2$ could present a more gradual switching if these parameters are modified, as evident in \autoref{fig:article2-fig1}b). In this work, we fixed the protocol for the three technologies to favor different device responses, which will be evaluated using the VDSP learning algorithm. 

Asymmetries in the histograms also reveal that LTP and LTD are not equivalent, and that HZO shows a more gradual RESET transition while CMO-HfO$_2$ shows a more gradual SET transition. This effect is captured by the steepness of the transition below/above the negative/positive threshold in the heat map representation.

\autoref{fig:article2-fig2}d presents the final weight reached after a write pulse programming event, and the histograms present the cumulative count of the final weight. TiO$_2$ shows an overall homogeneous distribution of weight achievable during the protocol, while HZO and CMO-HfO$_2$ show a more asymmetric distribution. HZO tends to reach the LRS more often, while CMO-HfO$_2$ tends to reach the HRS more frequently. This is consistent with the asymmetry in $\Delta W$ observed in \autoref{fig:article2-fig2}c.

\subsection{Device modeling}

\begin{figure}[H]
     \centering
         \includegraphics[clip,width=18cm,height=10cm,keepaspectratio, width=1\textwidth]{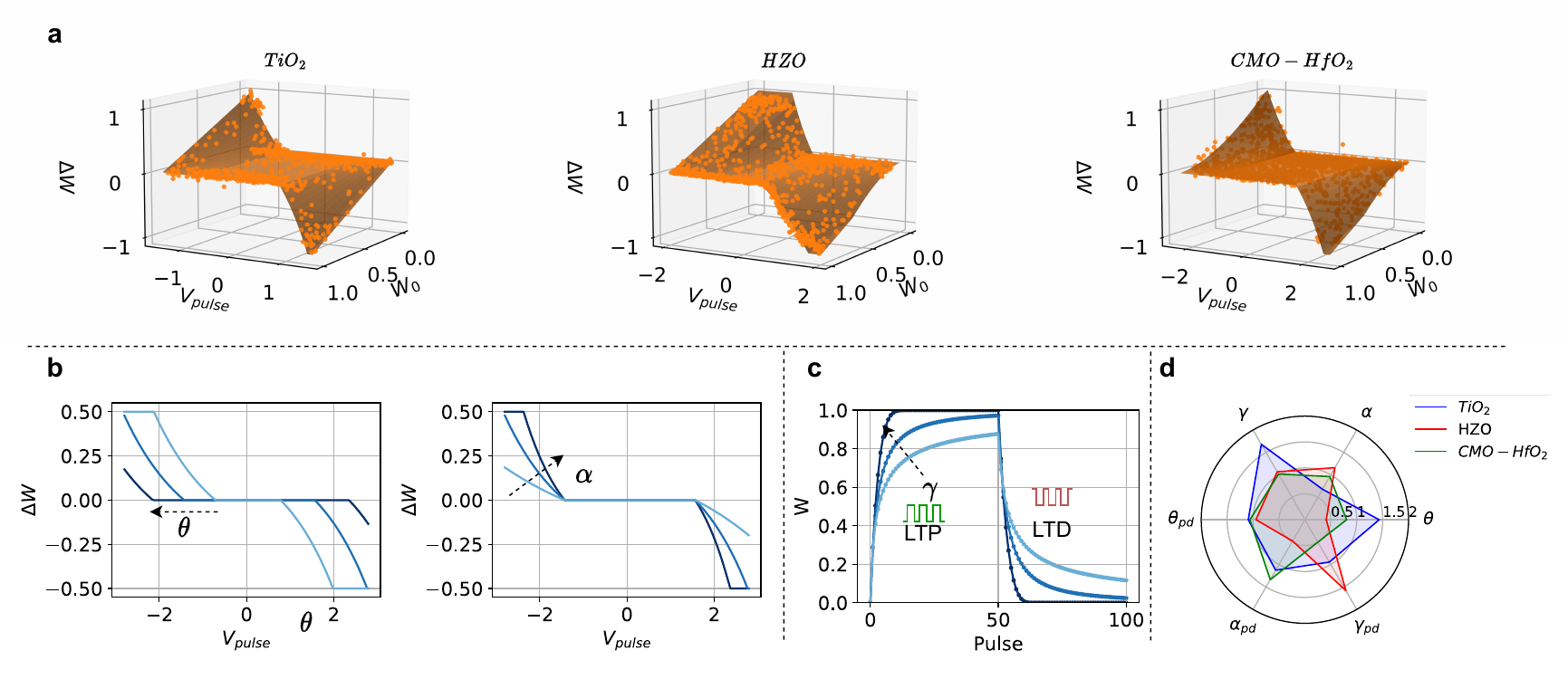}
    \caption{Model fitting.
    {\footnotesize
    \textbf{a} 3D visualization of the modeled $\Delta W$ is shown as a surface plot against the applied voltage ($V_{pulse}$) and the initial weight ($W_0$). Characterization points for the $TiO_2$, HZO, and CMO-HfO$_2$ devices are also presented. 
    \textbf{b} The impact of fitting parameters $\theta$ (left) and $\alpha$ (right) illustrates the variation in memristive switching threshold and curvature, respectively. Both plots show the change in weight as a result of a single voltage pulse, with $W_0 = 0.5$. 
    \textbf{c} Cumulative potentiation/depression plot obtained from 50 pulses of positive polarity (LTP) followed by pulses of negative polarity (LTD). The resultant curve of final weight with respect to three values of $\gamma$ is shown to illustrate the non-linearity fitting. 
    \textbf{d} Fitted model parameters for the three device stacks.
    }}

    \label{fig:article2-fig3}
\end{figure}

We subsequently model the voltage-driven modification of the memristor's weight for various initial conditions and programming pulse magnitude. The alteration in weight ($\Delta W$) is represented as the product of a switching rate function $f$ dependent on the applied voltage $v$ and a window function $g$ dependent on the weight $W$:

\begin{equation}
    \Delta W = f(v) \cdot g(W)
\end{equation}

\begin{equation}
    f(v) = \begin{cases} 
              e^{-\alpha_p \cdot (v - \theta_p)} - 1 & \text{if } v < \theta_p \\
              e^{\alpha_d \cdot (v - \theta_d)} - 1 & \text{if } v > \theta_d 
           \end{cases}
    \label{eq:f_of_v}
\end{equation}

Where $\alpha_p$ and $\alpha_d$ are exponential curvature fitting parameters for potentiation and depression, $v$ is the applied voltage, and $\theta_p$ and $\theta_d$ are the threshold voltages for memristive device switching for potentiation and depression, respectively. The window function \autoref{eq:g_of_W} describes the dependence of the weight change on the initial state ($w$) and is responsible for the multiplicative effect during cumulative switching events.

\begin{equation}
    g(W) = \begin{cases}
        (1 - w)^{\gamma_p} & \text{if } v < \theta_p \\
        w^{\gamma_d} & \text{if } v > \theta_d
    \end{cases}
    \label{eq:g_of_W}
\end{equation}

Where $\gamma_p$ and $\gamma_d$ are the non-linear fitting parameters for potentiation and depression. This non-linearity implies that a particular voltage pulse has a diminished impact on the device's conductance when applied multiple times. 

The model parameters for different resistive and ferroelectric memristive devices are compared in \autoref{fig:article2-fig3}d and presented in \autoref{tab:article2-table1}. The 3D representation with data points from the characterization and model is compared in \autoref{fig:article2-fig3}a to depict the model's accuracy (a 2D representation of model behavior on characterization points is shown in \suppautoref{fig:article2-figs1}). The model effectively captures device behavior and allows for quantitative metrics such as threshold, curvature, and state dependence. The effect of $\theta$ and $\alpha$ on $\Delta W$ is shown for $W_o = 0.5$ in \autoref{fig:article2-fig3}b. 

Next, to examine the state dependence, programming was performed using 50 LTP pulses of +1V followed by LTD pulses of -1V, as illustrated for three values of $\gamma$. Note that all the center lines in the three plots correspond to the fitted parameters for TiO$_2$. In \autoref{fig:article2-fig3}d and \autoref{tab:article2-table1}, all parameters ($\theta$, $\alpha$, $\gamma$) are compared for the TiO$_2$, HZO, and CMO-HfO$_2$ devices.

\begin{table}[h!]
\centering
\resizebox{\textwidth}{!}{%

\begin{tabular}{ccccccccccc}
\hline
\textbf{Device} & \textbf{$\alpha_p$} & \textbf{$\alpha_d$} & \textbf{$v_{thp}$} & \textbf{$v_{thd}$} & \textbf{$\gamma_p$} & \textbf{$\gamma_d$} & \textbf{HRS ($\Omega$)} & \textbf{LRS ($\Omega$)} & \textbf{RMSE} & \textbf{sf$_{pd}$} \\ \hline
TiO$_2$        & 0.678               & 0.762               & 1.432              & 1.563              & 1.68               & 1.583              & 15k                 & 2k                  & 0.047        & 1.057         \\ \hline
HZO            & 1.159               & 0.549               & 0.411              & 0.387              & 1.067              & 1.684              & 45M                 & 17M                 & 0.041        & 1.2           \\ \hline
CMO-HfO$_2$    & 0.96                & 1.27                & 0.8                & 0.85               & 1.017              & 0.5                & 4k                  & 1k                  & 0.0141       & 1             \\ \hline
\end{tabular}}
\caption{Model parameters for TiO$_2$, HZO, and CMO-HfO$_2$ devices along with the high resistance state (HRS) and low resistance state (LRS). Additionally, the table presents the root mean square error (RMSE) of $\Delta W$, comparing characterization data with model predictions.}
\label{tab:article2-table1}
\end{table}

In particular, the TiO$_2$ device has the highest $\theta$, implying a high switching threshold. This suggests a wide dead zone and a high SET/RESET voltage requirement. The $\alpha_p$ is, however, the lowest, implying a gradual dependence on weight change by increasing the SET voltage. The $\alpha_d$ is lower, pointing toward the fact that RESET is better controlled by voltage than SET. Finally, the $\gamma$, or state-dependent non-linearity, is highest in these devices. Previous studies have reported these behaviors \cite{seo2011analog}, and they can also be observed in \autoref{fig:article2-fig1}b.

HZO exhibits the lowest $\alpha_d$, implying that LTD is most gradual with respect to voltage (slightest curvature). The $\alpha_p$ is, however, twice the value of $\alpha_d$, signifying a strong asymmetry in voltage-dependent switching. Finally, $\gamma_d$ shows a strong asymmetry: state-dependent non-linearity is more evident in LTD. Such behaviors can also be observed in the IV sweep and LTP/LTD plots shown in \autoref{fig:article2-fig1}b, highlighting that the memristive device model strongly correlates with the known device behaviors. For CMO-HfO$_2$ devices, $\gamma$ is the lowest compared to the other two devices, signifying linear multi-level programming, which is also evident in \autoref{fig:article2-fig1}b. The $\alpha$ has a higher LTD value than LTP, similar to TiO$_2$: gradual weight change occurs with increasing reset voltage. $\gamma_d$ is smaller than $\gamma_p$, indicating that RESET is more linear than SET. However, it is important to note in this parametric model that different sets of parameters could explain the same switching response for a device. For instance, as per \autoref{eq:f_of_v}, $\alpha$ and $\theta$ have an inverse relationship, where a higher value of $\alpha$ can still fit the data points with a low error rate if $\theta$ is lowered accordingly.

\subsection{SNN benchmark}

\begin{figure}[H]
     \centering
         \includegraphics[clip,width=18cm,height=12cm,keepaspectratio, width=1\textwidth]{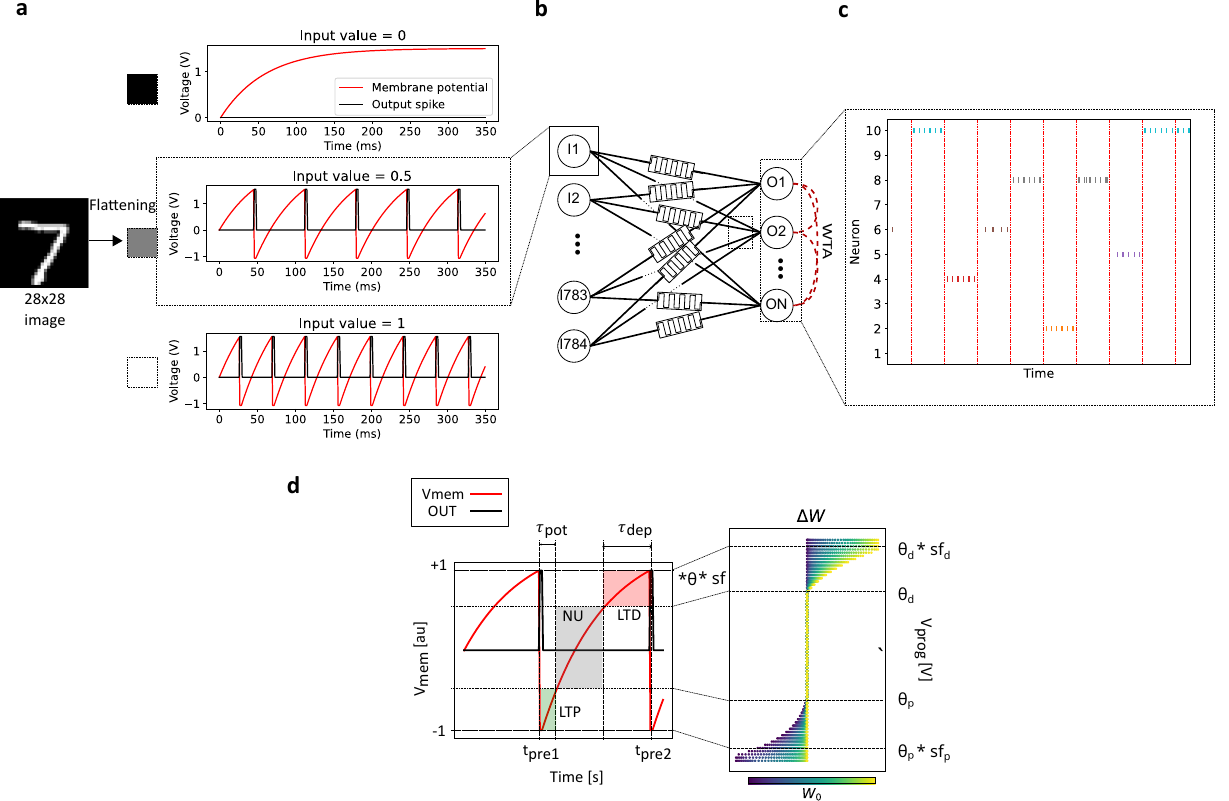}
    \caption{MNIST benchmark and simulation-device mapping in SNN.
    {\footnotesize
    \textbf{a} The input LIF neurons integrate a constant input current based on pixel values to implement rate-encoding.
    \textbf{b} Each neuron receives input from the individual pixels of a 28x28 image and is fully connected to $N$ output neurons through memristive synapses.
    \textbf{c} A raster plot for a network comprising 10 neurons illustrates the response to 10 sample inputs. The red-dotted vertical line distinguishes between different samples.
    \textbf{d} (left) The LIF neuron's membrane potential and spikes under constant stimulation input. (right) Weight change ($\Delta W$) versus applied voltage. The scaling factor (sf) and switching thresholds ($\theta$) are used to map the neuron membrane potential to the memristive device programming window. Long-term potentiation (LTP), long-term depression (LTD), and no update (NU) can be adapted to the memristive device requirements.
    }}

    \label{fig:article2-fig4}
\end{figure}

In order to assess learning capabilities, we conducted training on a Spiking Neural Network (SNN) using the MNIST dataset to recognize handwritten digits. In this method, we supplied the input pixels as constant currents to the encoding layer (refer to \autoref{fig:article2-fig4}a), where the Leaky Integrate-and-Fire (LIF) neurons convert the image's pixels into spike trains with frequencies proportional to their respective intensities. Additionally, we introduced Gaussian noise into the input pixels to induce stochastic sampling of the membrane potential (and thus programming voltage) during weight update events. This means that pixels receiving active input undergo different degrees of weight update magnitude. This enhances the realism of the evaluation by accounting for environmental noise and fluctuations in process temperature in encoding circuits, and also replicates the stochasticity observed in biological systems (Poisson distributed spikes). 

These spikes are weighted by the memristive synapse (\autoref{fig:article2-fig4}b) and integrated by neurons in the output layer. The neurons in the output layer are connected through a winner-take-all (WTA) topology, which leads to only one active output neuron at a given instance, corresponding to the network's decision, as shown in \autoref{fig:article2-fig4}c. A detailed description of the neuron model, training/evaluation procedure, and hyper-parameters is provided in the Methods section.

A scaling factor ($sf$) is used to tune the actual voltage applied to the memristor so that it matches the operational range of the memristive device. This factor essentially translates the computing signals (in the form of membrane potential) into the appropriate voltage levels that a memristor requires for switching. The relationship between these parameters can be expressed as follows:

\begin{equation}
    V_{prog} = Vmem \cdot sf \cdot \theta
    \label{eq:sf}
\end{equation}

In this equation, the programming voltage ($V_{prog}$) is the product of the neuron membrane potential ($V_{mem}$), the memristor's fitted threshold value ($\theta$), and the scaling factor ($sf$). \autoref{fig:article2-fig4}d captures the impact of the memristive threshold and scaling factor on the temporal response sensitivity of VDSP, i.e., the time window in which a post-synaptic spike could occur for the synaptic weight to change. In particular, if a postsynaptic spike occurs between $t_{pre1}$ and $t_{pre1} + \tau_{pot}$, the synapse will experience potentiation. On the other hand, a postsynaptic spike that occurs between $t_{pre2}$ and $t_{pre2} - \tau_{dep}$ will lead to depression. These areas of LTP and LTD depend on whether the programming voltage would be able to surpass the memristive switching threshold. Since the programming voltage depends on the membrane potential, $\theta$, and scaling factor (shown in \autoref{eq:sf}), the scaling factor directly influences the regions of LTP and LTD, thereby affecting the temporal sensitivity window of VDSP. For instance, a lower scaling factor would necessitate a higher membrane potential for the programming voltage to exceed the memristive switching threshold. As a result, the pre-neuron must be closer to firing, leading to a shorter time difference required between the pre- and post-neuron spikes to induce a weight change.

In the case of VDSP, the degree of change in weight in response to a single training sample depends on (i) the number of update instances, which depends on the spiking frequency of the output neuron, (ii) how many update instances fall into the LTP/LTD regions, and (iii) the magnitude of weight change defined by the fitted parameters of the memristive model. The first is tunable by changing sample presentation duration, leak rate, and the threshold of output neurons, i.e., the network hyper-parameters. The second and third depend on the scaling factor and the fitted parameters or physical switching characteristics of the memristive devices. Thus, it is essential to adjust the scaling factor for both the memristive device and the classification problem.

\begin{figure}[H]
     \centering
         \includegraphics[clip,width=18cm,height=10cm,keepaspectratio, width=1\textwidth]{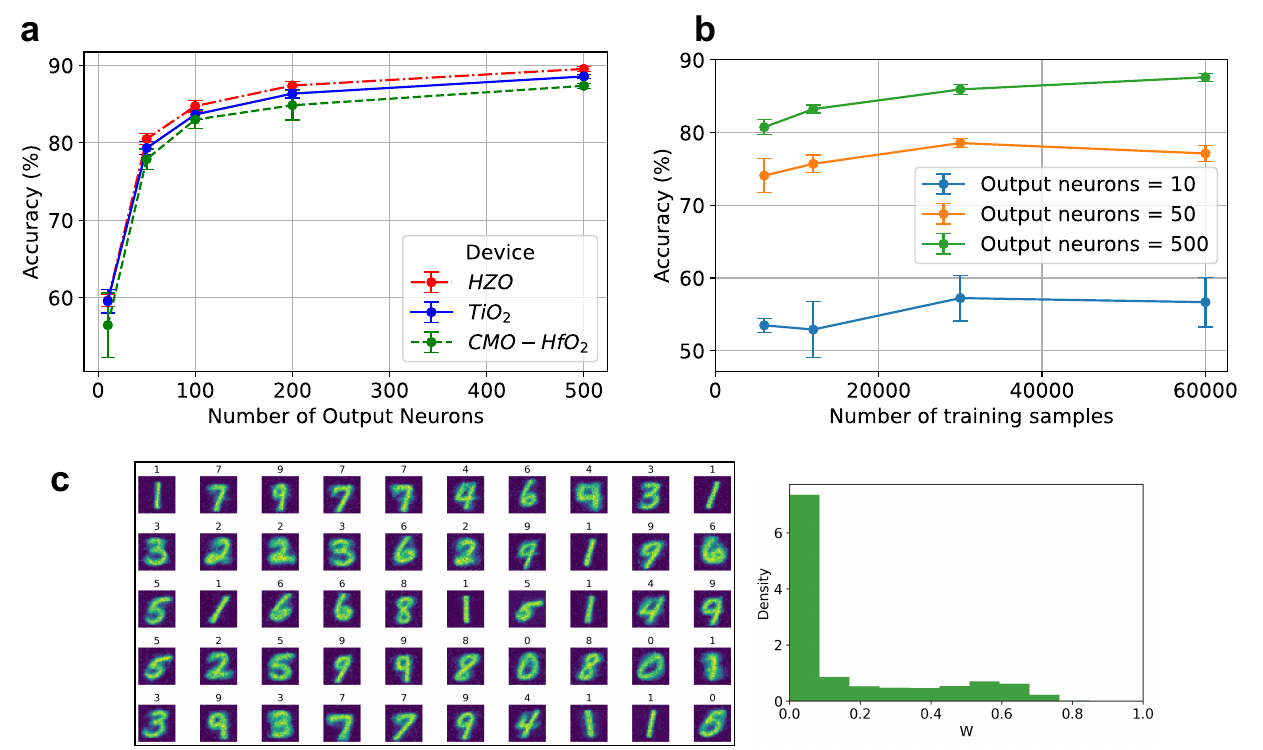}
    \caption{MNIST benchmark results.
    {\footnotesize
    \textbf{a} The test accuracy is shown as a function of the network size, indicated by the number of output neurons, for all three devices. The network was trained with three epochs of the MNIST dataset, and the average and standard deviation from five experiments with different initial weights are represented as error bars.
    \textbf{b} Performance evolution of the network based on the number of training samples used, for networks with 10, 50, and 500 neurons. The results correspond to the $TiO_2$ fitted model.
    \textbf{c} The weight plots and histogram after training are displayed for a network comprising 50 neurons.
    }}
    \label{fig:article2-fig5}
\end{figure}

The training subset of the MNIST database was used to perform unsupervised learning and evaluation for SNN for up to three epochs. \autoref{fig:article2-fig5}a illustrates the resulting performance for networks with 10 to 500 output neurons, trained with three epochs of 60,000 MNIST samples. For a network of 10 output neurons, the recognition rate was 60\%, which increased to more than 88\% for a network of 500 neurons. In addition, to evaluate incremental learning, a network of 10, 50, and 200 output neurons was trained up to a single epoch, and the labels were assigned by presenting 10,000 unseen digits from the remaining dataset. The experiments were carried out for five different initial conditions (weights), and the average and standard deviation are shown as error bars in \autoref{fig:article2-fig5}b. A similar analysis for incremental learning with HZO and CMO-HfO$_2$ device parameters is illustrated in \suppautoref{fig:article2-figs4} and \suppautoref{fig:article2-figs5}.

Larger networks allow for the learning of receptive field areas (inspired by neuron selectivity in neural cells \cite{bienenstock1982theory}) or distinct features for each class \cite{diehl2015unsupervised}. In other words, non-overlapping representations are essential to distinguish numbers with overlapping characteristics, such as the vertical line of one and nine. For example, the learned weights of a network of 50 output neurons are shown in \autoref{fig:article2-fig5}c, where complementary representations of each class can be seen. Competitive learning \cite{rumelhart1985feature} was implemented using the winner-take-all (WTA) mechanism in the neurons of the output layer to learn such complementary features. In addition, the histogram of the network’s weights at the end of training shows a bimodal distribution (\autoref{fig:article2-fig5}c), resulting from soft clipping due to the state-dependent multiplicative component of the VDSP update function. This clipping, a known non-linearity \cite{yang2022nonlinearity} of the memristive transition during repeated LTP/LTD programming, as shown in \autoref{fig:article2-fig1}b and is beneficial for online learning as it promotes stable learning without forgetting previously learned patterns~\cite{covi2016analog}.

\begin{table}[h]

    \centering
    \resizebox{\textwidth}{!}{%

    \begin{tabular}{cccccc}
        \hline
        Ref & Device & Circuit & Architecture & Plasticity & Accuracy \\ \hline
        \cite{boybat2018neuromorphic} & PCM & 8-R Multicell & 784x50 & STDP & 70\% \\ \hline
        \cite{nandakumar2018phase} & PCM & 2-R Differential & 784x350x10 & Supervised & 80\% \\ \hline
        \cite{srinivasan2016magnetic} & MTJ & 1-R & 784x[100]100 & Stochastic STDP & 70\% \\ \hline
        \cite{guo2019unsupervised} & HfOx/TaOy & 1T1R & 784x50 & STDP & 75\% \\ \hline
        \cite{roldan2022spiking} & 2D h-BN & 1R & 784x500 & STDP & 68\% \\ \hline
        \cite{shin2022pattern} & PCM & 6T2R & 724x500 & STDP & 73.6\% \\ \hline
        \cite{Querlioz2013ImmunityTD} & Ag/Si ECM & 1R & 784x50 & Simplified STDP & 80\% \\ \hline
        This work & TiO\textsubscript{2} & 1R & 784x50 & VDSP & \textbf{79\%} \\ \hline
        This work & $HZO$ & 1R & 784x50 & VDSP & \textbf{81\%} \\ \hline
        This work & CMO-HfO\textsubscript{2} & 1R & 784x50 & VDSP & \textbf{78\%} \\ \hline
    \end{tabular}}
    \caption{Comparison of the current study with previous memristive-based online learning benchmarks with MNIST. Different device technologies, including Phase change (PCM), Magnetic tunnel junction (MTJ), and electrochemical metallization (ECM) with circuit configurations, are tabulated for classic, stochastic, and simplified versions of STDP with respective network architectures.}
    \label{tab:article2-table2} 
\end{table}

\autoref{tab:article2-table2}\footnotemark compares the performance with previously reported works. Among the reviewed works, \cite{shin2022pattern} reported an energy consumption of 8.95 pJ per synapse, and \cite{srinivasan2016magnetic} reported 29.3 pJ per synapse, both estimated through circuit-level simulations. In all other studies, the programming pulses were generated using benchtop instruments, and details on circuit-level implementations, area, and energy overhead were not provided. In our VDSP approach, by decoupling the time window for spike correlations in Hebbian learning from the programming pulse width, we achieved pulse durations between 200 ns and 1 \textmu s, highlighting its practical efficiency. The area of the neural building block with a 16×16 network was 4 mm², as reported in \cite{garg2024neuromorphic}. Among the devices investigated in this study, HZO exhibited the highest performance, followed by TiO$_2$ and CMO-HfO$_2$. Overall, the learning efficiency of VDSP for all three devices was equivalent to or superior to that of their STDP counterparts. All the network parameters, such as the degree of noise, leak rate of input neurons, threshold, and output neurons, were optimized through the TiO$_2$ device model for a network of 10 output neurons. Only the scaling factor for LTP and LTD was optimized through grid search ($sf_p$, $sf_{pd}$) for the three devices and number of samples in the incremental training experiment, with the optimized values plotted in \suppautoref{fig:article2-figs2} and \suppautoref{fig:article2-figs2b} . The $sf$ plays a similar role to the learning rate in ANNs as it regulates the degree of weight change. 

It is essential to highlight that, unlike classical machine learning optimizations and gradient-based learning in ANNs, the notion of learning rate is not trivial in SNNs, specifically in cases of unsupervised local online learning where there is no batch processing. The magnitude of the weight change depends on the difference in spike times between the presynaptic and postsynaptic neurons. The choice of learning rate is critical to avoid local minima or continuous oscillations, as a sub-optimal rate slows learning and hinders convergence. On the other hand, a higher learning rate can cause instability in weights, lead to forgetting previously learned information, and make convergence difficult.

\footnotetext{All results are obtained from software-based simulations of MNIST classification.}

\subsection{Impact of device variations}

\begin{figure}[H]
     \centering
         \includegraphics[clip,width=18cm,height=10cm,keepaspectratio, width=1\textwidth]{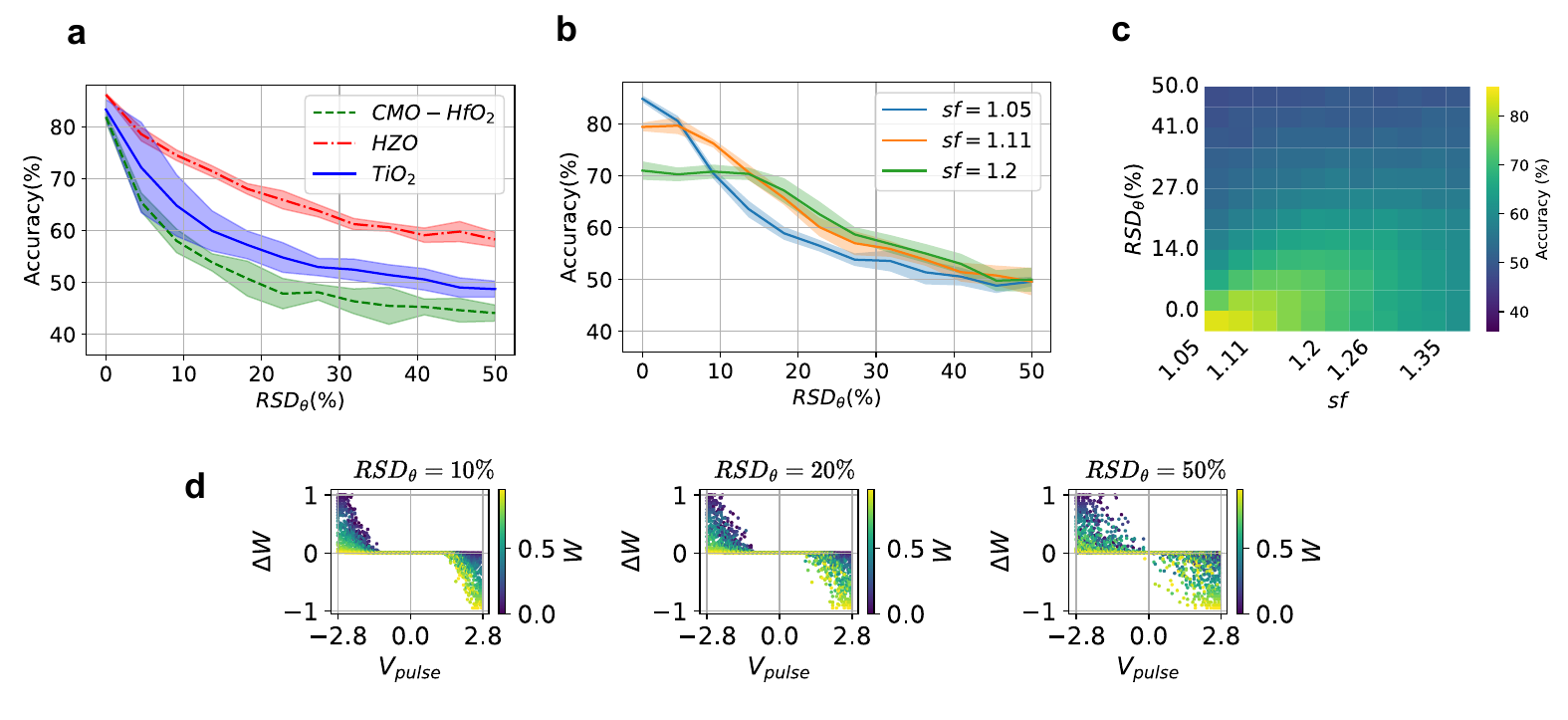}
    \caption{Impact of device-to-device variability in switching threshold.
    {\footnotesize
    \textbf{a} The switching thresholds ($\theta$) for the 784x200 synapses are sampled from a normal distribution centered on the fitted model parameter. The relative standard dispersion, represented as $\frac{\sigma}{\mu}$, varies and is noted as $RSD_\theta$.
    \textbf{b} The resulting accuracy is shown for three distinct values of $sf$ for $TiO_2$ device parameters. (Note that the other two devices follow a similar trend, see \suppautoref{fig:article2-figs4} and \suppautoref{fig:article2-figs5}).
    \textbf{c} A detailed grid search of $sf$ and $RSD_\theta$ was conducted, and the resulting average accuracy over ten experiments is displayed as a 2-D heatmap.
    \textbf{d} Plots of device characteristics depicting $\Delta W$ versus $V_{pulse}$ to demonstrate the variability effects for three $\frac{\sigma}{\mu}$ levels: 0.1, 0.2, and 0.5 (arranged from left to right).
    }}

    \label{fig:article2-fig6}
\end{figure}

The SET/RESET voltage or the switching threshold of memristive devices varies from device to device, particularly in the case of analog switching. The corresponding memristor model parameter ($\theta$) is sampled from a normal distribution, with the mean centered around the fitted parameter and a relative standard deviation (RSD) that was incrementally changed to study its impact on the device’s behavior. RSD is defined as the ratio of the standard deviation to the arithmetic mean of a normal distribution ($\frac{\sigma}{\mu}$). A histogram of sampled threshold distribution for different values of RSD is shown in \suppautoref{fig:article2-figs3}. \autoref{fig:article2-fig6}a illustrates how network performance is affected by different degrees of threshold variability for a neural network with 200 output neurons using model parameters from three different devices. As the RSD increases, the performance of the network degrades. For the TiO$_2$ device, with 20\% variability in the switching threshold, the performance of 82\% drops to 56\%. The corresponding fitted parameter ($\theta$) threshold of these devices was 1.4V and 1.5V for LTP and LTD, respectively. A similar analysis was conducted for the HZO device (see \suppautoref{fig:article2-figs4}), and its performance dropped to 68\%. These FTJs have the lowest thresholds, around 0.4V, but the highest $\alpha$ or curvature. The achievable conductance range in ferroelectric devices is strongly correlated with the magnitude of the programming voltage. Therefore, a greater scaling factor could be used, as detailed in \suppautoref{fig:article2-figs2}. The low switching threshold makes STDP implementation with pulse overlapping challenging. The pulse overlapping technique requires transmitting spikes through sub-threshold pulses for non-destructive reading; thus, the maximum programming voltage for LTP/LTD is limited to twice this threshold.

The scaling factor is a crucial parameter that determines the probability of switching in memory devices. A baseline value of 1.05 prevents devices with a higher threshold from switching, limiting the number of devices that can synapse to learn. To accommodate threshold variations and increase the number of such devices, a larger scaling factor should be used. In \autoref{fig:article2-fig6}b, it is demonstrated that a high scaling factor of 1.2 keeps performance stable, with only a slight decrease from 71\% to 68\% when the variability is 20\%. On the other hand, a scaling factor of 1.05 achieves a performance of over 82\% without considering device mismatch. However, when variability is introduced, the performance drops to less than 60\%. This highlights the importance of selecting the right scaling factor to maintain strong performance, especially in the presence of variability.

The use of a high scaling factor can help reduce performance drops caused by variability, but it can also hinder generalized learning that utilizes the entire dataset. This creates a trade-off between stability and learning precision. A high scaling factor increases the magnitude of the programming voltage, leading to a greater weight change magnitude for each update. However, when training with a dataset like MNIST, the goal is to generalize over all the training samples and learn incrementally from each sample. Therefore, it is crucial to regulate the impact of each training sample on the weights, which can be achieved by reducing the scaling factor. A parametric sweep between RSD and sf, shown in \autoref{fig:article2-fig6}c, further illustrates how different scaling factors impact device performance. Additionally, \autoref{fig:article2-fig6}d plots the influence of threshold mismatch on the switching function of the devices for different variabilities in switching parameters. These results are based on the TiO$_2$ model parameters, with similar trends observed for the other two device stacks (CMO-HfO$_2$ and HZO), as shown in \suppautoref{fig:article2-figs4} and \suppautoref{fig:article2-figs5}. This consistency across different materials highlights a generalizable behavior of the scaling factor for accommodating the threshold mismatches across multiple resistive memory devices. In summary, even with 50\% variability, the performance across all three devices remains above 45\%.

\begin{figure}[H]
     \centering
         \includegraphics[clip,width=18cm,height=10cm,keepaspectratio, width=1\textwidth]{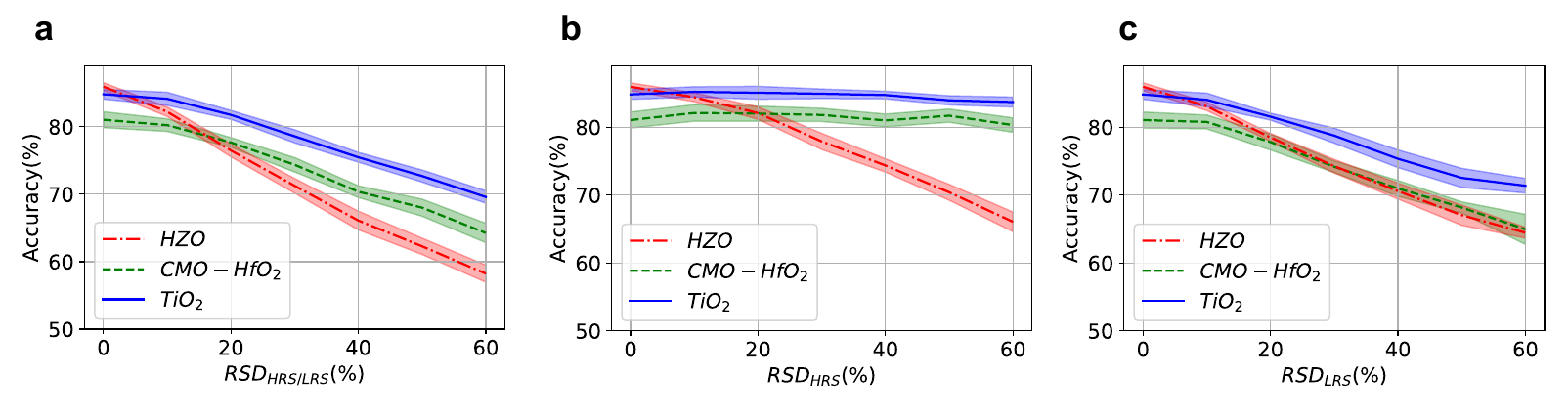}
   \caption{Impact of device-to-device variability in HRS and LRS levels.
    {\footnotesize
    \textbf{a} The HRS and LRS for the 784x200 synapses are sampled from a normal distribution, centered on the fitted model parameter. The relative standard dispersion, represented as $\frac{\sigma}{\mu}$, varies and is noted as $RSD_\theta$. The resulting accuracy is shown for three devices.
    \textbf{b} The impact of HRS is isolated and evaluated.
    \textbf{c} The impact of LRS is isolated and evaluated. Each experiment was repeated with ten different initial conditions, with the lines and shades depicting the mean and standard deviation of measured accuracy.
    }}

    \label{fig:article2-fig7}
\end{figure}

The impact of variations in both the low resistance state (LRS) and high resistance state (HRS) on a 784x200 synaptic network was analyzed (\autoref{fig:article2-fig7}). Device-to-device variability was simulated by sampling the HRS and LRS of each synapse from a normal distribution centered around the measured device values. The standard deviation (RSD\textsubscript{HRS} or RSD\textsubscript{LRS}) of these distributions was adjusted to investigate its effect on the network’s recognition accuracy. Both HRS and LRS were sampled simultaneously from their respective distributions (\autoref{fig:article2-fig7}a). To further differentiate the effects of HRS and LRS individually, \autoref{fig:article2-fig7}b presents the impact of variability in HRS, while \autoref{fig:article2-fig7}c presents the effects of LRS mismatch (RSD\textsubscript{LRS}) on the network’s performance.

In \autoref{fig:article2-fig7}a, HZO devices achieve the highest accuracy despite having the lowest HRS/LRS ratio among the three devices investigated, indicating that under ideal conditions without variability, performance is largely independent of this ratio. Here, HRS and LRS denote the extreme boundaries of the device’s analog conductance. Each fractional weight state is assumed to shift proportionally within these new boundaries when device variability is introduced. However, device-to-device variations cause a more pronounced accuracy drop in HZO. As shown in \autoref{fig:article2-fig7}b, even under extreme HRS/LRS variability, TiO\textsubscript{2} and CMO-HfO\textsubscript{2} devices retain their original accuracy, while HZO experiences a performance decline. Thus, although a higher HRS/LRS ratio helps mitigate variability, it is not a critical parameter for learning in the absence of such variations.

TiO\textsubscript{2}-based memories exhibit the highest resilience to variations in high and low resistance states (\autoref{fig:article2-fig7}a), followed by CMO-HfO\textsubscript{2} and HZO. This trend aligns with their measured ON/OFF ratios from \autoref{tab:article2-table1}, where TiO\textsubscript{2} shows the highest ratio (7), while CMO-HfO\textsubscript{2} and HZO measure 4 and 3, respectively. Because TiO\textsubscript{2} and CMO-HfO\textsubscript{2} have larger ON/OFF ratios, changes in the HRS state (w=0) have minimal impact on subsequent layer activity, reducing the performance sensitivity (\autoref{fig:article2-fig7}b). However, when HRS values approach LRS, the effect on the activity of the output neuron can become significant, particularly in devices with smaller resistance ranges. \autoref{fig:article2-fig7}c further shows that all three devices are equally affected by LRS variations, but even with a variation of 20\%, the accuracy drop remains below 10\%, underscoring the overall robustness of the system. Consequently, LRS variability is less impactful than HRS variability, especially in devices with larger resistance ranges.

\section{Discussion}

\subsection{Device-Specific Switching Dynamics}

There are two types of switching that we observe. The first type involves a pulse of the same magnitude and pulse width that causes the device to transition between different conductance states. This is also known as cumulative switching, and it is exploited by STDP for online learning. The transition is non-linear, and the magnitude of weight change reduces when moving to the boundary. The second mechanism shows that the conductance state strongly depends on the switching voltage or pulse width. By adjusting the programming voltage level, we can expand the boundaries of switching voltage. The three devices exhibit different degrees of cumulative or voltage-dependent switching in LTP or LTD, depending on the dynamics of underlying mechanisms like oxidation, filament rupture, or ferroelectric domain switching. Some processes are self-limiting, allowing robust control, while others are not and lead to abrupt behaviors. For instance, (i) in TiO$_2$, the SET process involves slower drift of oxygen vacancies~\cite{zhao2011dynamic} and is more gradual or cumulative in comparison to the RESET process, which involves conductive filament melting. These resistive devices often experience noisy switching due to variability in oxygen vacancy migration~\cite{ryu2020artificial, lee2012change}. They have a high ON/OFF ratio, so the learning is resilient to variations in the ratio. (ii) In HZO-based FTJs, the SET process tends to be more abrupt compared to the RESET process, which is typically gradual and exhibits better linearity~\cite{ryu2019ferroelectric}. (iii) CMO-HfO$_2$ devices often display variability in resistance states (HRS and LRS) due to random oxygen vacancy movements, and exhibits more granularity in potentiation over depression~\cite{sassine2016interfacial}.  VDSP-based learning effectively mitigates differences in device parameters, leading to similar MNIST test-set recognition rates despite the distinct parameter combinations exhibited by the devices due to different underlying physical mechanisms of switching. Scaling factors and the asymmetry in the scaling factor can be tuned with respect to the device threshold and asymmetry in switching dynamics. This scaling factor is an important determinant of the network's learning rate.  There exists a trade-off between gradual weight updates with a low scaling factor and abrupt updates with a higher scaling factor. While a high scaling factor improves resilience to variations in the device’s switching threshold, gradual updates help the model generalize better by learning incrementally from each sample.  The effective number of stable analog states is governed by the match between the programming pulse profile and the device’s switching dynamics. Because the conductance levels in nonlinear analog devices are unevenly spaced, a precise state count is difficult; nevertheless, this non-uniformity, while suboptimal for conventional digital storage, naturally regularizes weight updates and imposes soft bounds in online learning for spiking neural networks. A detailed discussion is provided in \suppautoref{fig:article2-figs9}

\subsection{Programming Strategy and Variability}

Resistive memories exhibit significant variability when scaled down due to the resolution limits of semiconductor patterning techniques and variations in fabrication parameters. This variability causes each device to exhibit slight differences in performance. These variations pose particular challenges in analog computing, complicating precise and accurate programming. In the case of SNNs with memristive weights, this variability directly affects the synaptic learning process, as different devices may exhibit different switching thresholds and ON/OFF states. This inconsistency makes it difficult to reliably program the desired resistance states across all devices, ultimately impacting the overall performance and accuracy of the network. One approach is to map the device's switching characteristics onto the learning algorithm, including its threshold voltages and asymmetric response to programming pulses. This requires selecting an appropriate learning rate to adjust synaptic weight updates, compensating for device variations. Fine-tuning network hyperparameters, such as the scaling factor that links neuron membrane potential to memristive programming voltage, ensures robust learning. By adapting these scaling factors to account for different switching thresholds, we show that the network can maintain reliable performance despite device variability. This robustness is driven by two key factors: the application of online learning and the advantages provided by VDSP. Firstly, it is well-established that the challenges posed by variability and noise in memristive devices can be mitigated through online learning strategies~\cite{yu2011stochastic}. Through continual adaptation, online learning enables the system to overcome fluctuations and inconsistencies inherent to the hardware. Device behaviors like gradual switching, which allows for fine-grained adjustments to synaptic weights, can mitigate the effects of variability. Additionally, devices with a broader range of conductance states can support more precise weight changes, reducing the impact of any single variation. For instance, memristors with cumulative switching, where the conductance increases gradually with repeated pulses, tend to be more resilient to slight variations in pulse width or amplitude, providing a built-in mechanism for error correction. Secondly, and perhaps more crucially, VDSP principles enhance resilience against device mismatch. In traditional STDP implementations that use pulse overlapping techniques, the amplitude and width of the programming pulses must be carefully adjusted based on the characteristics of the synaptic device. Furthermore, the programming pulse width is tailored to meet the specific requirements of the Hebbian learning window, ensuring that spike correlations occur within the window to induce synaptic changes. However, VDSP simplifies this process by utilizing a continuous physical quantity—namely, the neuron membrane potential—amplified and directly applied for synaptic programming. A higher-than-necessary amplification factor ensures that even mismatched memristors with high switching thresholds are programmed successfully. Additionally, noise originating from the input layer—whether from the encoding circuit or environmental sources—contributes to the membrane potential and, consequently, the programming voltage. This added noise, combined with variability among memristors, introduces a stochastic element into the learning process. This randomness allows for finer weight adjustments, which can smooth out inconsistencies due to device mismatch and improve learning accuracy.

The degradation of memristors with time is a critical phenomenon for currently available devices. In this regard, we have studied the impact of drift on phase-change memories (PCM) in another study~\cite{palhares202428, quintino202428} where we discussed the robustness of the learned receptive fields through VDSP \cite{garg2022voltage} to such drift at room and cryogenic temperature and proposed circuits to mitigate the same. Online learning can partially compensate for slow drift through continual updates, as shown in related works in the domain of non-spiking neural networks \cite{antolini2023combined}.  

Another non-ideality to consider is the limited lifetime of such memory devices or the number of programming cycles the device could sustain, often called endurance. With repeated programming cycles, the ON/OFF ratio of the device degrades. VDSP uses short pulses that should increase the device lifetime by avoiding over-stress in conventional pulse overlapping techniques. Nevertheless, in earlier sections, we show that the learning is resilient to the absolute value of the ON/OFF ratio. Variations in the switching threshold are also essential to evaluate the impact of device degradation, which makes some devices stuck to a specific level. With a variability of 20\% in theta and sf=1.2, more than 40\% of devices are stuck (\suppautoref{fig:article2-figs3}). However, as shown in \autoref{fig:article2-fig6}b, the accuracy is not impacted significantly. This indicates the overall tolerance to stuck devices due to extreme degradation. Additional circuit-level solutions (e.g., periodic refresh or calibration pulses) and advanced device engineering would be necessary for long-term reliability to mitigate wear-out and retention loss.

In this study, we focus on two types of device stacks: valence change memory (VCM) devices based on TiO\textsubscript{2} and CMO-HfO\textsubscript{2}, and ferroelectric devices based on HZO. However, other devices, such as electrochemical metallization (ECM) devices and magnetic tunnel junctions (MTJs), may exhibit greater stochasticity in their switching behavior \cite{suri2013bio}. We anticipate that VDSP can be adapted similarly to the probabilistic spike-timing-dependent plasticity \cite{ambrogio2016unsupervised}, where the neuron membrane voltage (Vmem) would determine the switching probability. Therefore, future research should aim to quantify the effects of this stochastic behavior and explore additional material stacks. 

\subsection{Hardware Considerations}

One of the key motivations behind the proposition of VDSP is the overall reduction in the hardware overhead. Although STDP can achieve similar MNIST accuracies, it often requires larger local memory to track pre- and post-synaptic timing variables on hardware or complex pulse shaping for implementation with memristors \cite{wang20152, ishii2019chip}. In contrast, VDSP folds the timing dependence into the neuron’s membrane voltage, eliminating overlapping pulses, and this reduces hardware complexity in principle. In addition, VDSP improves the robustness to variations in input signal frequencies. To illustrate the reduced hardware overhead of VDSP compared to STDP, simplified hyperparameters, and robustness to variations in input signal frequencies, we performed a set of control simulations in which input frequencies ranged from 5Hz to 500Hz. While STDP’s learning efficiency diminished outside its optimal spike-timing window, VDSP maintained \>90\% of its maximal accuracy across this frequency range, as shown in \cite{garg2022voltage}. We also compare the number of local parameters: VDSP requires only membrane voltage and spike signals, whereas STDP-based synapses typically store pre- and post-synaptic traces, increasing per-synapse complexity by up to 3 times as the temporal learning window parameters (A+/A- and Tau+/Tau-) are also configurable learning parameters set according to the dynamics of the input signal.  

In a typical crossbar architecture, each pre-synaptic neuron generates a programming pulse amplitude proportional to its membrane potential. An example of a CMOS integrated selector-based crossbar implementation to switch between inference and weight update phases is illustrated in \suppautoref{fig:article2-figs6}. This approach aligns well with conventional CMOS flows, as shown by prior 1T1R CMOS integrated memristive circuits \cite{shen2012high, shulaker2014monolithic, wan2022compute}. The main distinction is that VDSP avoids storing pre- and post-synaptic spike traces, thereby reducing local memory and pulse-shaping circuits. In another study, we proposed and validated memristor-interfacing circuits and analog spiking neurons to implement VDSP \cite{garg2024versatile}. 

In \suppautoref{fig:article2-figs7}, \suppautoref{fig:article2-figs8}, and \cite{garg2024neuromorphic}, we presented on-chip circuits and simulation results to amplify the membrane potential to match the memristor programming range, corresponding to the hardware implementation of the scaling factor (sf). STDP, on the other hand, requires a current conveyor circuit to ensure spike transmission and perform weight updates in parallel, thereby increasing the width of the transmitted spikes and, consequently, the energy consumption. In this study, we demonstrated that one needs only to adjust the scaling factor for LTP and LTD, which sets the amplitude of the programming pulse relative to the device threshold. Other hyperparameters (such as leak constant, threshold voltage, refractory period, etc.) remain consistent between TiO\textsubscript{2}, HZO, and CMO-HfO\textsubscript{2}. This choice demonstrates that VDSP can adapt to different devices without re-optimizing the entire training or neuron configuration, thus making our approach more hardware-friendly.

In the proposed architecture, the weight of the active post-neuron is updated serially. Such updates could also occur in parallel if the presynaptic neuron is connected to the memristive terminal instead of the gate of transistor in the 1T1R cell. However, parallel updates may present challenges, as updating the weight of multiple metal oxide memristive devices in parallel could result in significant current sourcing and sinking for the peripheral CMOS circuits. In serial implementation, a weight update pulse of 200 ns and a select / unselect time of 1 µs leads to 1.2 µs for updating a single device. Since the post-neuron has a refractory period on the order of milliseconds, more than 800 devices can be updated within this timeframe without inducing additional latency. Moreover, larger crossbars face limitations such as IR drop and sneak paths; therefore, recent studies \cite{payvand2022self} promote tile-based architectures with smaller crossbars to scale synaptic arrays.
\subsection{Scalibility}

In this study, we employ a minimal two-layer SNN primarily for clarity, allowing us to isolate device-level parameters and highlight their impact on system-level learning rather than optimize for state-of-the-art accuracy. This setup facilitates direct benchmarking and comparison of three device technologies with minimal confounding factors. Moreover, such a two-layer network also allows visualization of receptive fields learned by the network to evaluate the overall training qualitatively. Although relying on a single hidden layer means that transformations such as rotations or scaling are not learned, our results confirm that VDSP-based learning is feasible across diverse memristive devices with only minor parameter adjustments.  

The proposed VDSP learning rule and its STDP counterpart offer an alternative to gradient-descent-based supervised methods that are hardware-friendly but are examined here primarily in two-layer architectures. Transitioning to deeper networks introduces additional considerations, including hierarchical feature extraction through the introduction of a third factor \cite{halvagal2023combination, fremaux2016neuromodulated}. Nevertheless, our memristor-friendly approach to weight optimization in a two-layer network provides a valuable foundation for extending these methods to deeper, multilayer systems. Future efforts could explore translating this rule to different topologies such as convolutional \cite{goupy2023unsupervised}, recurrent, and multilayer networks, paving the way for improved accuracy and extraction of transformation-invariant features.

\section{Conclusion}

This article addresses designing and tuning computing circuits based on memristive device characteristics and demonstrates neuron state-based online learning with VDSP. This method is local in space and time, as it requires the application of the instantaneous analog membrane potential of the pre-synaptic neuron. The learning process does not require complex pulse-shaping circuitry and models the classical exponential dependence of weight change on the difference in spike time between pre- and post-synaptic neurons.  The slow neuron dynamics enable the memristive device to adapt its conductance based on the frequency and timing of spikes, responding to changes in membrane potential. This coupling is essential for enabling long-term potentiation and depression (LTP and LTD) in a bio-realistic manner. By combining the slow dynamics of neurons (voltage over time) with the voltage-dependent switching of memristors, we can achieve learning with a bio-realistic time scale using sub-microsecond programming pulses. Using short pulses saves power, increases endurance, and improves learning due to controlled switching. Limiting the pulse width reduces the energy consumed during each programming event, particularly in large-scale neuromorphic systems where energy efficiency is critical. Shorter pulses also decrease wear on the devices, enhance their endurance, and offer more precise control over weight updates. We characterized and modeled two resistive devices and a ferroelectric memristive device, each exhibiting distinctive switching behaviors. The characterization of voltage-dependent switching behavior was achieved by applying pulses of random voltage magnitude. Different devices exhibit properties such as switching threshold, non-linearity, and variability. A generalized model was proposed to describe resistive switching as dependent on the magnitude of the applied voltage and the resistance state of the device, with the fitted model closely resembling the characteristics of the tested devices. The model parameters accounted for fundamental memristive properties such as threshold, non-linearity, and asymmetry, enabling the validation of learning efficiency across a spectrum of device behavior deviations. Moreover, immunity to deviations in model parameters was observed by sampling the device model parameters from a distribution with varying spreads. Importantly, performance deterioration due to variability can be mitigated by tuning the scaling factor.

The proposed plasticity, implemented with simple CMOS circuits~\cite{garg2024versatile}, allows large-scale systems with limited energy and space constraints to carry out online learning for real-world pattern recognition applications. In future work, we plan to expand this approach to more complex patterns by using current-limited switching through 1T1R devices and investigating the relationship between the device's material stack and the resulting intermediate conductance states. The VDSP-based programming provides high-resolution memristive learning, effectively linking neuron dynamics with memristive properties to detect and respond to long-term and temporal patterns. However, because model parameters are heavily influenced by programming conditions, attributing performance solely to the device stack is speculative, and further research is required to fully understand these interactions.

\section{Methods}
\subsection{Device Fabrication}

The TiO$_{2-x}$ cross point device is created at the intersection between a TiN bottom electrode and the remaining Al$_2$O$_3$/TiO$_x$/Ti/TiN stack. The Damascene process was used to fabricate an embedded electrode and minimize the surface topography and roughness. First, 400~nm deep trenches defined by e-beam lithography were etched into a Si$_3$N$_4$ passivation layer and then filled by 600~nm of TiN deposited by reactive sputtering. Chemical mechanical polishing was used to remove the excess TiN between the lines and planarize the surface. The rest of the stack was then deposited starting from a 1.5~nm Al$_2$O$_3$ tunnel barrier using ALD, followed by the 30~nm sub-stoichiometric TiO$_{2-x}$ switching layer, 10~nm Ti oxygen reservoir and 30~nm TiN inert electrode, deposited by reactive sputtering (TiO$_{2-x}$) and sputtering (Ti, TiN) without breaking the vacuum. To reduce the top electrode resistance, a 400~nm Al layer was also deposited on top of the active stack using e-beam evaporation. The top electrode and active stack lines were defined by electron-beam lithography and etched by Cl-based plasma. Finally, PECVD SiO$_2$ was used to encapsulate the device, and the vias in SiO$_2$ were opened to expose the bottom and top electrode pads. More information on the fabrication process is available in~\cite{el2022fully}.

For the fabrication of the HZO-based devices, a thermal oxidation process was used to grow a 200~nm thick SiO$_2$ layer on a silicon substrate. The active layers were subsequently deposited using plasma-enhanced atomic layer deposition (PE-ALD). Initially, a 20~nm TiN bottom electrode was formed at 300\,$^\circ$C. This was followed by the deposition of a 2~nm WO$_x$ layer at 375\,$^\circ$, and then a 3.5~nm thick HZO layer was grown at 300\,$^\circ$C. An additional 10~nm TiN capping layer was deposited on top. To induce crystallization, millisecond flash lamp annealing was employed. The sample was preheated to 400\,$^\circ$C and then subjected to a 20~ms energy pulse delivering 90~J\,cm$^{-2}$. Subsequently, a 100~nm thick tungsten (W) top electrode was deposited by sputtering. The junction area was defined through optical lithography followed by reactive ion etching (RIE) of the W and upper TiN layers, with the HZO layer serving as an etch stop. The bottom electrode was patterned using optical lithography and etched using inductively coupled plasma (ICP) RIE through the HZO, WO$_x$, and TiN layers. A 100~nm SiO$_2$ passivation layer was deposited at 300\,$^\circ$C by plasma-enhanced chemical vapor deposition (PECVD). Vias to access both top and bottom electrodes were opened using optical lithography. SiO$_2$ was etched with RIE, and the underlying HZO and WO$_x$ layers were subsequently etched using ICP to expose the TiN contact. This step was directly followed by sputtering of 100~nm of W. The first-level metal interconnects were defined using lithography and RIE. A second 100~nm SiO$_2$ passivation layer was then deposited by PECVD at 300\,$^\circ$C. Vias to the bottom contacts were patterned using lithography and etched in the SiO$_2$ layer using RIE. A final 100~nm W layer was sputtered, and the second-level metal lines were defined using standard lithography and etching techniques. More details on the fabrication process are available in ~\cite{begon2022scaled}.

CMO-HfO\textsubscript{2} based device was fabricated using CMOS- and BEOL-compatible process flows. The bottom electrode consists of a 20 nm TiN layer deposited by Plasma-Enhanced Atomic Layer Deposition (PEALD) at 300\textsuperscript{0}C, followed by the formation of a 6 nm HfO\textsubscript{}{2} switching layer using the same technique. The top stack—comprising 50 nm of W, 20 nm of TiN, and 20 nm of CMO—was deposited by sputtering and subsequently encapsulated with a 100 nm SiN passivation layer. A final 100 nm W top electrode was also added by sputtering. Additional process-related information is available in \cite{falcone2023physical}.

All three devices are fabricated with CMOS compatible processes, allowing back-end-of-line integration with transistor circuitry \cite{el2022fully, begon2024back}. Integration of CMO/HfO\textsubscript{2} devices has already been demonstrated by \cite{falcone2025all},  and similar integration for the other two device stacks is ongoing (\suppautoref{fig:article2-figs11}).

\subsection{Characterization setup}

Electrical measurements were performed on an Agilent B1500A semiconductor analyzer and with a B1530A waveform generator/fast measurement unit (WGFMU). Write pulses were generated by a remote-sense and switch unit (RSU) module close to the probe and applied to the top electrode while the bottom electrode was grounded. The resistance of the device was measured at V = +/-100 mV with a high resolution source measurement unit (SMU) on the top electrode, while the bottom electrode was grounded.

\subsection{Model fitting}

The Levenberg-Marquardt least squares fitting algorithm (lmfit) \cite{newville_matthew_2014_11813} was used to fit the model parameters to the characterization data. 

\subsection{SNN simulations}

Leaky Integrate-and-Fire (LIF) neurons~\cite{abbott1999lapicque}, are simplified models of biological neurons, making them efficient to simulate within a SNN simulator. This neuron model was used for the presynaptic neuron layers. The corresponding equation is
\begin{equation}
    \tau_m \frac{dv}{dt} = -v + I + b
    \label{eq1}
\end{equation}

where $\tau _{m}$ denotes the membrane leak time constant, $v$ represents the membrane potential, which decays to the resting potential ($v_{rest}$), $I$ is the injected current, and $b$ is a bias term. When the membrane potential exceeds a threshold level ($v_{th}$), the neuron fires a spike. Subsequently, it becomes unresponsive to any input during the refractory period ($t_{ref}$) and the neuron potential is reset to the voltage ($v_{reset}$).

In the output layer, an adaptive leaky integrate and fire (ALIF) neuron model was used, which has an additional state variable $n$, which increases by $inc_n$ with each spike, and its value is deduced from the input current. This leads the neuron to decrease its firing rate over time when exposed to high input currents \cite{la2004minimal}. The state variable $n$ decays with a time constant $\tau _{adap}$ :

\begin{equation}
    \tau _{adap} \frac{dn}{dt} = -n
    \label{eq2}
\end{equation}

\begin{table}[H]
\begin{tabular}{|l|l|l|l|}
\hline
Parameter      & Input layer & Output layer & Description                      \\ \hline
$v_{th}$            & 1           & 8            & Spiking threshold                \\ \hline
$\tau _{mem}$            & 30ms        & 12ms         & Leak time constant               \\ \hline
$t_{ref}$           & 3ms         & 3ms          & Refractory period                \\ \hline
$inc_n$       & -           & 1            & Threshold increment              \\ \hline
$\tau _{adap}$ & -           & 120ms        & Adaptive threshold leak          \\ \hline
$t_{wta}$         & -           & 12ms         & Lateral inhibition time constant \\ \hline
\end{tabular}
\caption{SNN simulation parameters for the MNIST benchmark}
\label{tab:table3}
\end{table}

The above parameters were optimized for a network of 10 output neurons and kept constant for different devices and network sizes. The initial optimization was performed through the Bayesian optimization preset in the Optuna library \cite{akiba2019optuna}. 

Gaussian noise was used to induce stochasticity in the input layer, which creates a jitter in pixels and samples the learned features at each output spike. Bias current was applied to background pixels to penalize inactive pixels and is essential for regularization and preventing one neuron not to learn all digits (equal amount of de-potentiation of nonactive pixels for learning distinguishing features). Each image was presented for 40 ms, resulting in at most 3 spikes in the input layer per sample. Hard Winner takes all based lateral inhibition was applied in output layer, in which, all neurons were inhibited for a period of $\tau _{wta}$ on firing event of any neuron. The network parameters were tuned using genetic search with Optuna~\cite{akiba2019optuna} package for 1000 experiments for the parameters of the TiO$_2$ device. Afterwards, all network parameters were the same for all network sizes, epochs, and three devices. The training was performed without using labels, and at the end of training, the weights were fixed, and the last 10,000 samples were used for assigning the class to each output neuron. Subsequently, the samples from the MNIST dataset's test set were used to evaluate the recognition rate. 

\section*{Data Availability}

Publicly available datasets were analyzed in this study, with this data available at http://yann.lecun.com/exdb/mnist/. All codes supporting this study can be accessed at https://github.com/nikhil-garg/VDSP-Memristors. 

\section*{Code Availability}

All codes supporting this study can be accessed at https://github.com/nikhil-garg/VDSP-Memristors. 

\section*{Acknowledgments}

We want to acknowledge engineers from 3IT, Sherbrooke, for TiO$_2$ device fabrication. The authors acknowledge the staff of Binnig and Rohrer Nanotechnology Center (BRNC), Zurich, for HZO and CMO-HfO$_2$ device fabrication. 

\section*{Funding}

We acknowledged financial supports from the EU: ERC-2017-COG project IONOS (\# GA 773228) and CHIST-ERA UNICO project. This work was also supported by the Natural Sciences and Engineering Research Council of Canada (NSERC) (No. 559730) and Fond de Recherche du Québec Nature et Technologies (FRQNT). FA thanks the MEIE for support through the chair in neuromorphic engineering.






\section*{Author Contributions Statement}

J.H., N.G. formulated characterization protocol. J.H., L.B.L., and Da.F. performed device characterizations. N.G. devised and fitted the model. N.G. and I.B. implemented and performed the SNN simulations. L.B., V.B., T.S., and Do.F.F. performed device fabrication. F.A., D.D., Y.B., D.Q., J.M.P., and B.J.O. supervised and administered the current study. N.G. and F.A. wrote the initial draft of the manuscript. All authors provided critical feedback and helped shape the research, analysis, and manuscript.

\section*{Competing Interests Statement}
The authors declare no competing interests.

\renewcommand{\figurename}{Supplementary Fig.} 
\setcounter{figure}{0}


\printbibliography

\newpage

\setcounter{page}{1}

\setcounter{section}{0}

\section*{Supplementary Information}

\section{Model fitting}

\begin{figure}[H]
     \centering
         \includegraphics[clip, width=0.9\textwidth]{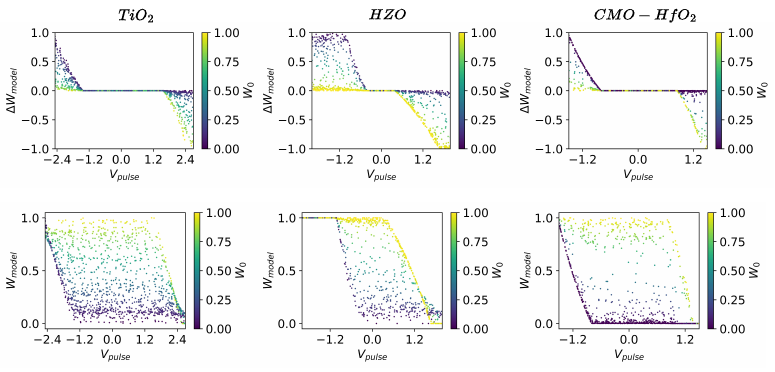}
\caption{The fitted model's prediction on characterization data points displays the weight change ($\Delta W$) and final weight ($W_f$) in relation to the applied voltage ($V_{pulse}$) and initial weight ($W_0$) for $TiO_2$, HZO, and CMO-HfO$_2$ measurement points (from left to right).}
    \label{fig:article2-figs1}
\end{figure}

\section{Optimal scaling factors}

\begin{figure}[H]
     \centering
         \includegraphics[clip,width=18cm,height=10cm,keepaspectratio, width=0.9\textwidth]{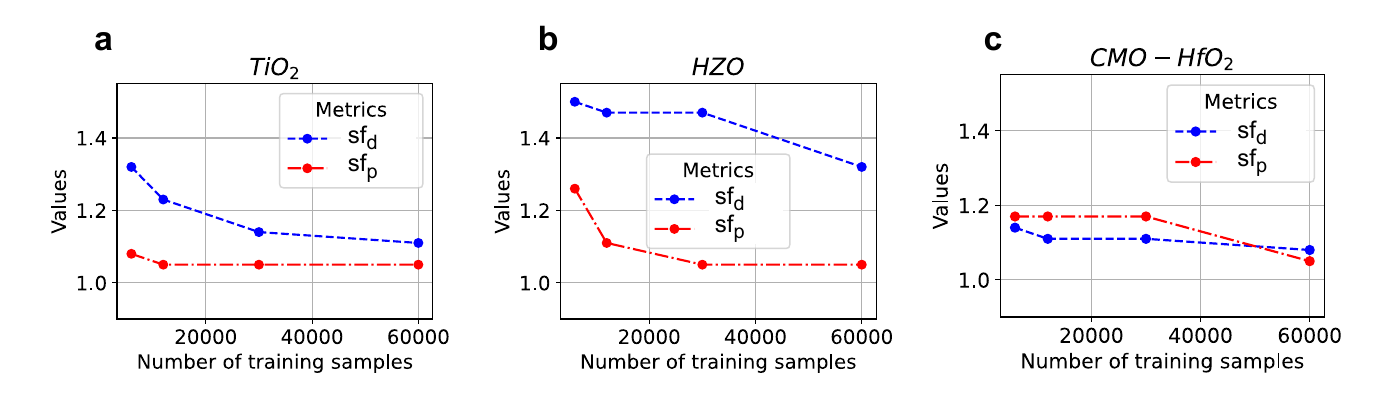}
    \caption{Scaling factor based on the number of training samples. The optimal scaling factors for LTP ($sf_p$) and LTD ($sf_d$) are depicted as a function of the number of training samples for a network consisting of 500 output neurons. The plots are presented for three devices: $TiO_2$, HZO, and CMO-HfO$_2$ device parameters (from left to right).
    }

    \label{fig:article2-figs2}
\end{figure}

\begin{figure}[H]
     \centering
         \includegraphics[clip,width=18cm,height=10cm,keepaspectratio, width=0.9\textwidth]{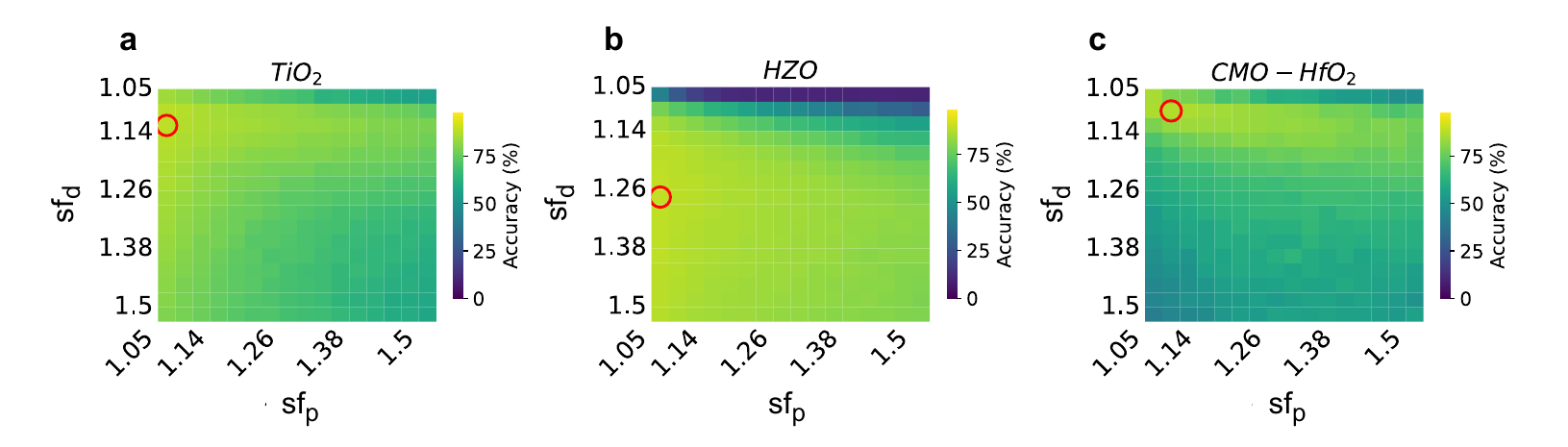}
    \caption{Impact of scaling factor on the classification performance of a network with 500 neurons. The test accuracy on the MNIST dataset is evaluated for different scaling factors for LTP ($sf_p$) and LTD ($sf_d$) in a network consisting of 500 output neurons. The plots are presented for three devices: $TiO_2$ (a), HZO (b), and CMO-HfO$_2$ (c) device parameters. The optimal combination of scaling factors that yields the maximum classification accuracy is encircled in red.
    }

    \label{fig:article2-figs2b}
\end{figure}

\section{Variability modeling}

\begin{figure}[H]
     \centering
         \includegraphics[clip,width=18cm,height=10cm,keepaspectratio, width=0.3\textwidth]{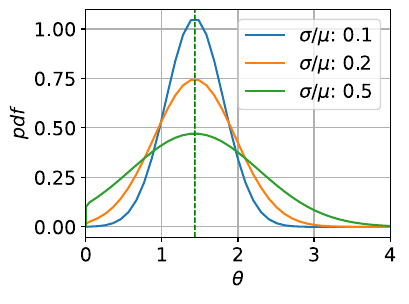}
    \caption{Probability distribution function of $\theta$ for $TiO_2$, showing variability in the form of relative standard dispersion ($\frac{\sigma}{\mu}$).
    }

    \label{fig:article2-figs3}
\end{figure}

\section{SNN benchmark analysis for HZO and CMO-HfO$_{2}$ devices}

\begin{figure}[H]
     \centering
         \includegraphics[clip,width=18cm,height=10cm,keepaspectratio, width=1\textwidth]{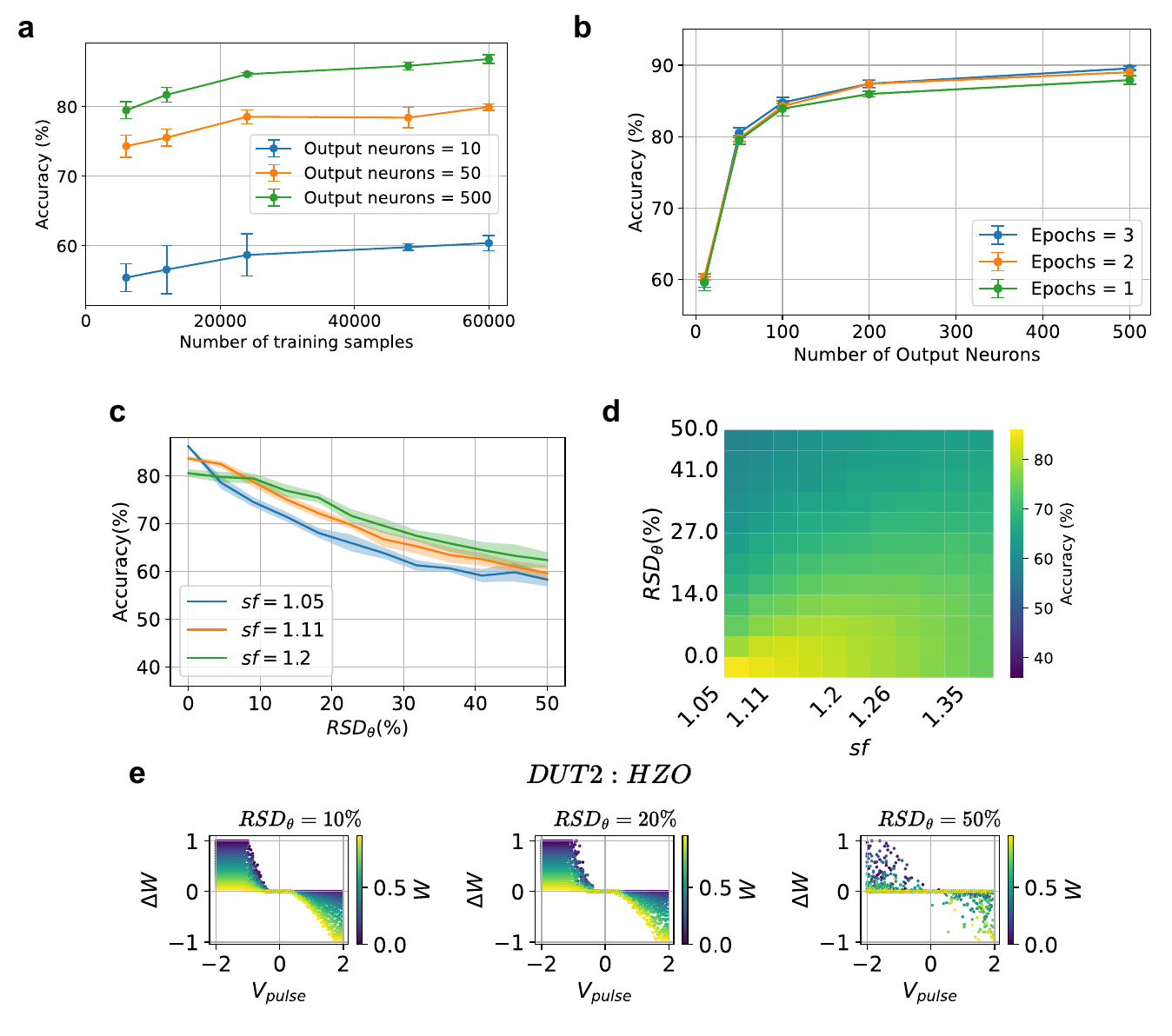}
    \caption{MNIST benchmark results for HZO device.
    {\footnotesize
    \textbf{a} Evolution of test performance in response to iterative training through examples from the MNIST dataset for 10, 50, and 500 output neurons.
    \textbf{b} Dependence of testing accuracy on the number of output neurons and training epochs. Each experiment was repeated five times with different initial weights, and the error bars show the standard deviation.
    \textbf{c} Impact of variability in switching threshold ($\theta$) plotted for three different scaling factors.
    \textbf{d} A detailed grid search of $sf$ and $RSD_\theta$ was conducted, and the resulting average accuracy over ten experiments is displayed as a 2-D heatmap.
    \textbf{e} Impact of varying degrees of variation in $\theta$ on the device switching characteristics.
    }}

    \label{fig:article2-figs4}
\end{figure}

\begin{figure}[H]
     \centering
         \includegraphics[clip,width=18cm,height=10cm,keepaspectratio, width=1\textwidth]{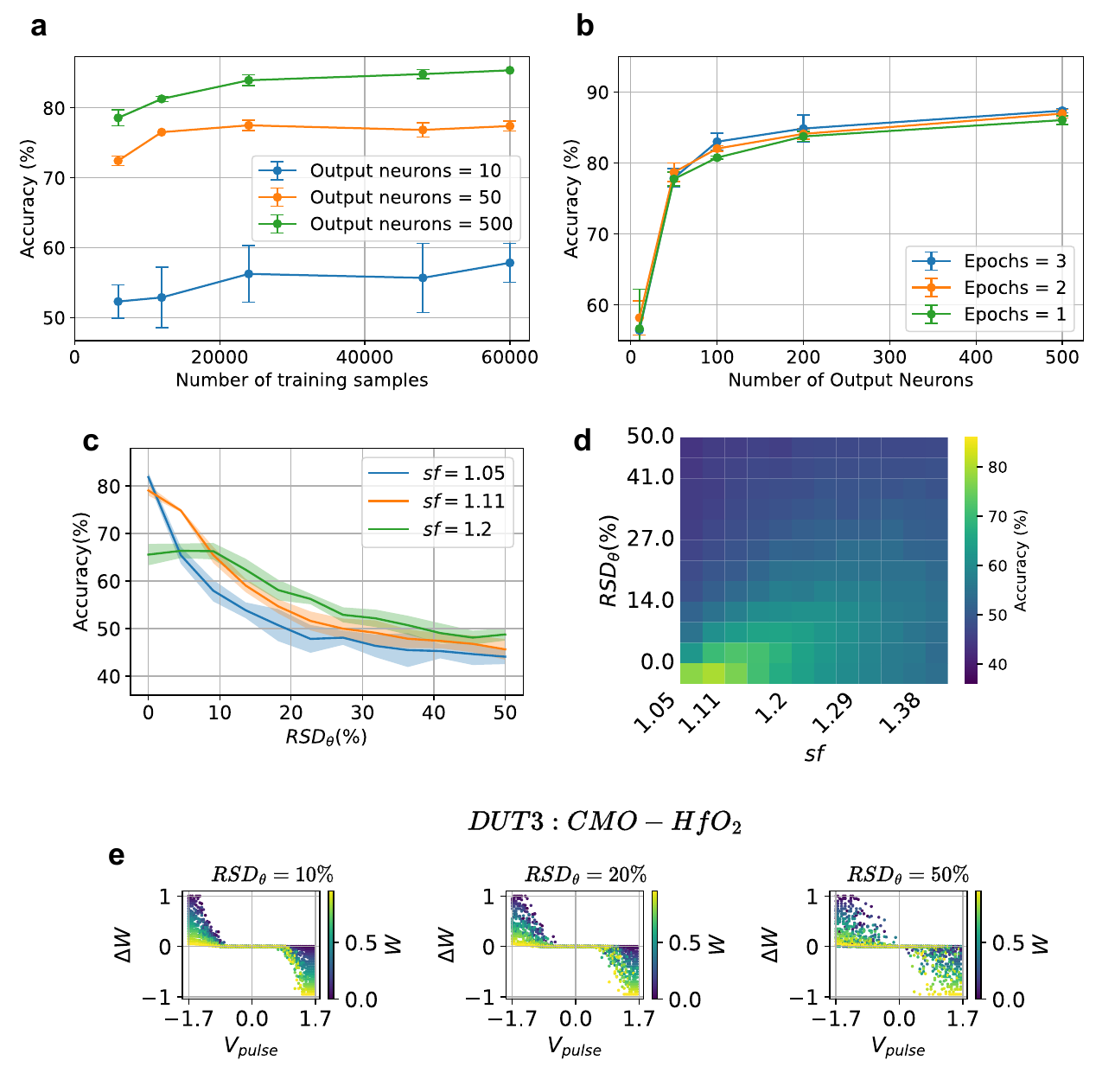}
    \caption{MNIST benchmark results for CMO-HfO$_2$ device.
    {\footnotesize
    \textbf{a} Test performance progression observed during iterative training using MNIST dataset examples with 10, 50, and 500 output neurons.
    \textbf{b} Relationship between testing accuracy and the number of output neurons and training epochs. Each experiment was repeated five times with different initial weights, with error bars indicating the standard deviation.
    \textbf{c} The effect of variability in the switching threshold ($\theta$) plotted for three distinct scaling factors.
    \textbf{d} A detailed grid search of $sf$ and $RSD_\theta$ was conducted, and the resulting average accuracy over ten experiments is displayed as a 2-D heatmap.
    \textbf{e} Effect of varying the degree of $\theta$ variation on device switching characteristics.
    }}

    \label{fig:article2-figs5}
\end{figure}

\section{Crossbar architecture and circuits}

\begin{figure}[H]
     \centering
         \includegraphics[clip,width=18cm,height=10cm,keepaspectratio, width=1\textwidth]{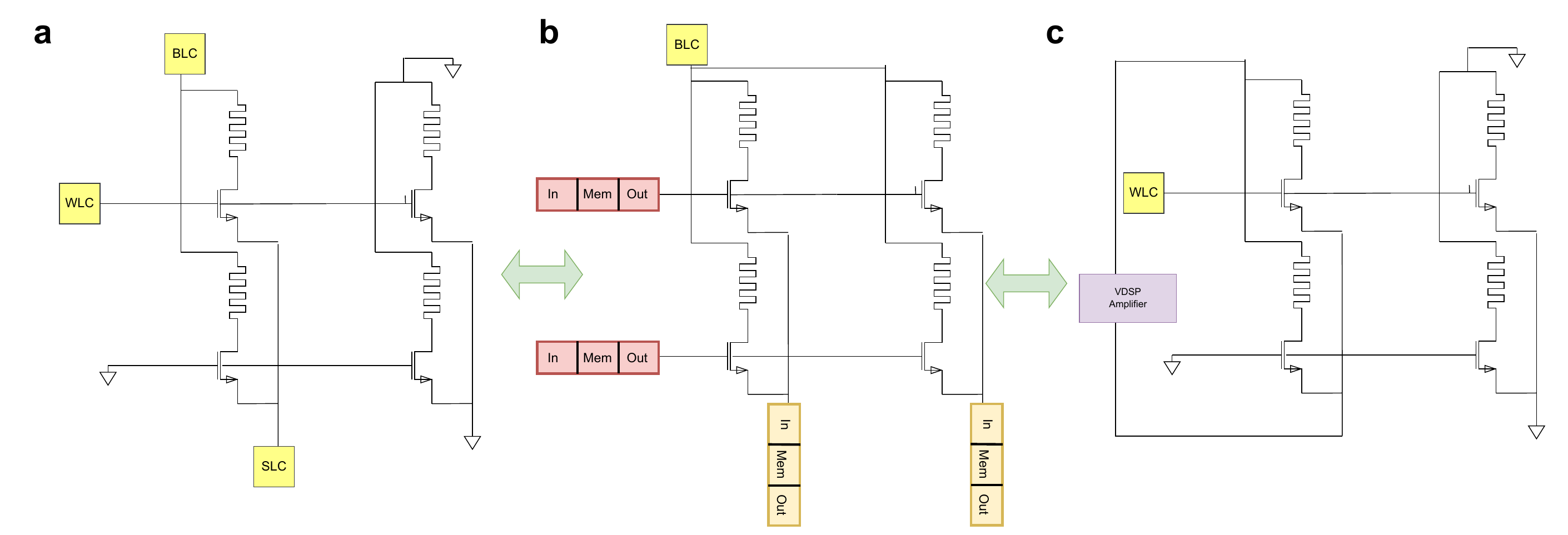}
    \caption{Crossbar architecture and modes of operations.
    {\footnotesize
    \textbf{a} In characterization mode, the IO pads (BLC/WLC/SLC) address the selected crossbar cell for forming, reading, and writing operations. \textbf{b} In inference mode, the output spike from the presynaptic LIF neuron is fed to the gate (WL) of the synaptic cell. The cell’s source is connected to the resistance reading and LIF blocks in the postsynaptic neuron bank. \textbf{c} In learning mode, a single synaptic cell is programmed by connecting the BL and SL to the on-chip amplifier, while the corresponding row is connected to the WLC IO pad to provide compliance current during the programming operation (externally).  
    }}
    \label{fig:article2-figs6}
\end{figure}

The crossbar operates in three distinct modes: characterization, inference, and learning. In the characterization phase (\suppautoref{fig:article2-figs6}b), a single 1T1R cell is selected and accessed via the bit line, word line, and source line characterization pads (BLC/WLC/SLC). This configuration is crucial for memristor forming, programming, and reading the state of individual cells. In the inference mode (\suppautoref{fig:article2-figs6}c), the synapses and neurons are interconnected to compute the network’s decision in response to inputs provided to the presynaptic neuron. The output spikes from the presynaptic neurons are applied to the gate of the respective 1T1R cells, while the source line is connected to the postsynaptic neuron input terminal. Lastly, during the learning phase (\suppautoref{fig:article2-figs6}d), the on-chip amplifier supplies the programming voltage to the bit line (BL) and source line (SL). The compliance current for programming is set externally by applying an appropriate voltage to the word line characterization pad (WLC). 

The architecture is designed with two key motivations: 
\begin{enumerate}
    \item Neuron-gate connection to the 1T1R cell: In this architecture, the neuron's output is connected to the gate of the 1T1R cell. When a neuron spike event occurs, it activates all memory cells in that row. The neuron only draws a small amount of current from each connected memory cell, while most of the read current is pulled from the bit line. This configuration eases the load on the neuron’s output block, enhancing the system’s scalability. 
    \item Efficiency in charging: The bit line (BL) remains continuously charged, and only the word line (WL) is charged during spike transmission. This provides both energy and latency benefits, improving overall system efficiency
\end{enumerate}

\section*{Circuits to convert membrane potential to a bi-polar programming voltage}

\begin{figure}[H]
     \centering
         \includegraphics[clip,width=18cm,height=10cm,keepaspectratio, width=1\textwidth]{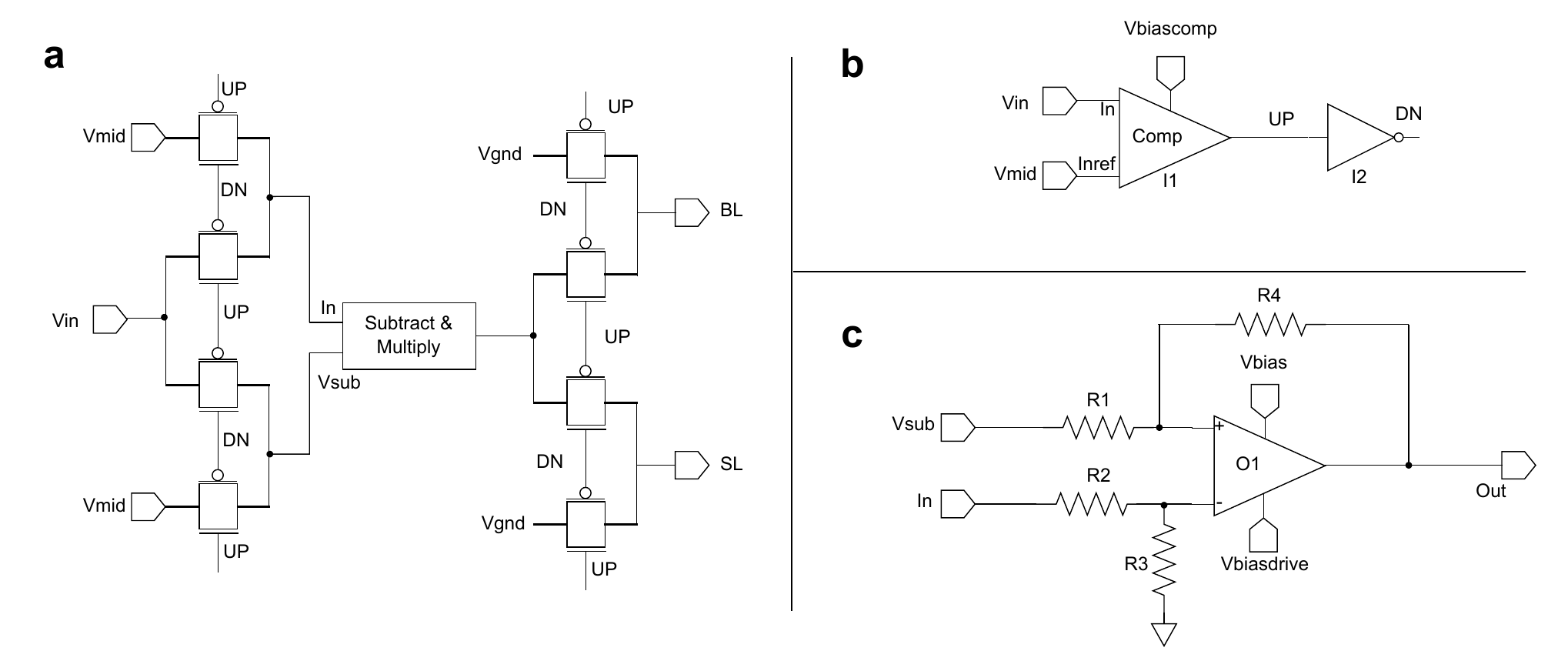}
    \caption{Circuits in the programming block (VDSP amplifier)
    {\footnotesize
    \textbf{a} Top-level schematic of the programming block consists of a subtract-and-multiply block and multiple transmission gates (TGATEs). \textbf{b} On-chip comparator circuit. \textbf{c} The subtract-and-multiply block. 
    }}
    \label{fig:article2-figs7}
\end{figure}

In the circuit-level implementation, the neuron membrane voltage remains strictly positive, as described by \cite{garg2024neuromorphic}. Because this operating voltage is lower than required to program the memristive synapses, supplementary interface circuits were introduced (\suppautoref{fig:article2-figs7}). These circuits route the unipolar membrane voltage to the synaptic cell's top or bottom terminal, depending on whether the membrane potential is above or below its resting value. The same interface stage amplifies the signal with an operational amplifier configuration that employs feedback resistors to set the desired gain or scaling factor (sf), as illustrated in \suppautoref{fig:article2-figs7}c.

\begin{figure}[H]
     \centering
         \includegraphics[clip,width=18cm,height=10cm,keepaspectratio, width=0.6\textwidth]{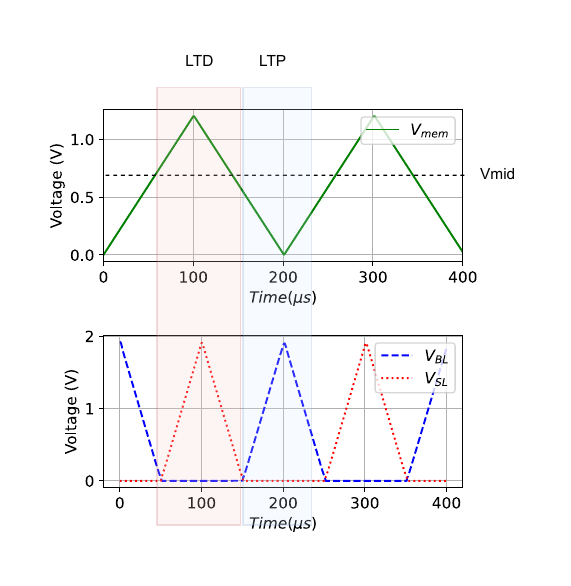}
    \caption{Response of the VDSP amplifier circuit in the programming block (obtained through SPICE simulations). The Vmem and Vmid (threshold for LTP/LTD) signals are applied as inputs, generating the VBL and VSL signals. 
    }
    \label{fig:article2-figs8}
\end{figure}

\suppautoref{fig:article2-figs8} presents SPICE simulation results for the programming and amplification circuits. The neuron‑membrane voltage Vmem ranges from 0 to 1.2 V, with the resting level Vmid fixed at 0.6 V. The resulting voltages applied to the top (bit‑line, BL) and bottom (source‑line, SL) electrodes are shown, delineating regions of potentiation (LTP) and depression (LTD). With the amplification factor set to 3, the programming voltage spans 0 – 1.8 V. 

\section{Analog switching}

\begin{figure}[H]
     \centering
         \includegraphics[clip,width=18cm,height=10cm,keepaspectratio, width=1\textwidth]{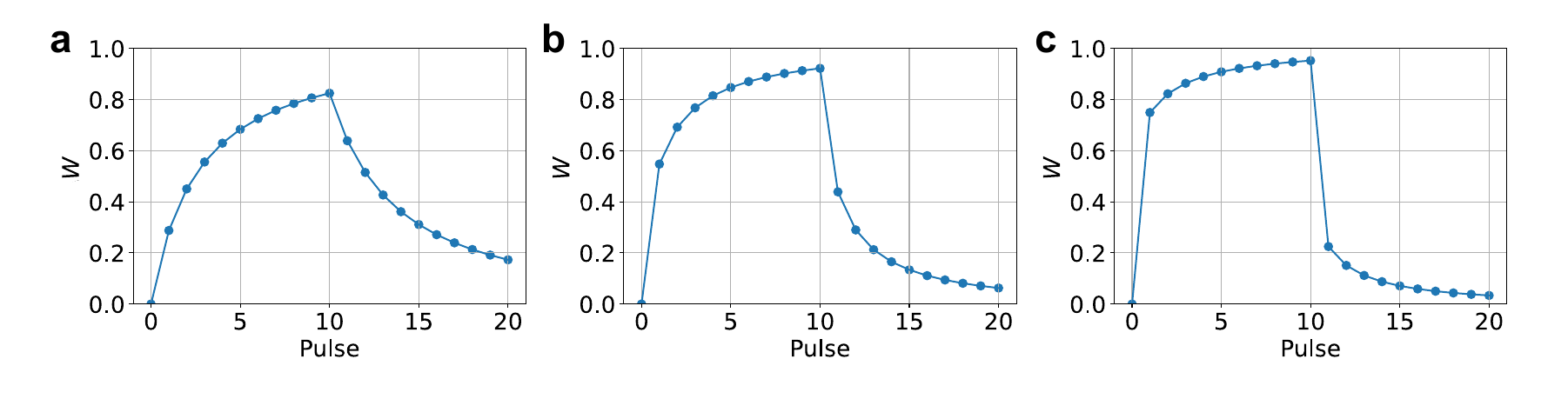}
    \caption{Number of intermediate conductance levels as a function of programming voltage, increasing from left to right (a-c). For each voltage, ten identical programming pulses induce potentiation, followed by ten pulses of opposite polarity to cause depression.}
    \label{fig:article2-figs9}
\end{figure}

The devices studied here switch in an analog regime. \suppautoref{fig:article2-figs9} highlights the switching non‑linearity: as the programming voltage increases, the number of accessible conductance levels shrinks. At the highest voltage, for instance, the first pulse changes the weight from 0 to 0.7, whereas nine additional pulses are needed to move from 0.7 to 1.0. Although such behavior is undesirable for traditional data‑storage applications or backpropagation implementation, it is well‑suited to online learning in SNNs. The weight‑dependent non‑linearity naturally regularizes updates and imposes “soft bounds” on the weights through the device physics. 

Furthermore, lowering the programming voltage restores gradual switching at the expense of a narrower conductance window (0.2 – 0.8). By exploiting the voltage‑modulated programming of VDSP, we can, therefore, combine the benefits of fine‑grained updates and a wide dynamic range simply by modulating the programming voltage. 

\begin{figure}[H]
     \centering
         \includegraphics[clip,width=18cm,height=10cm,keepaspectratio, width=1\textwidth]{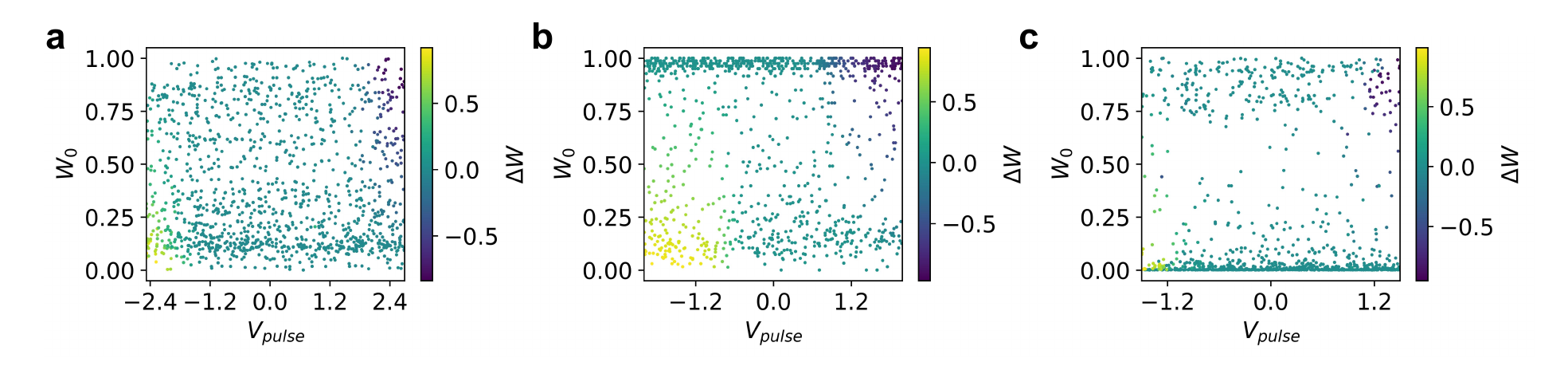}
    \caption{Electrical characterization results visualized as a scatter plot to show the dependence of change in weight (in color bar) as a function of applied voltage and initial weight for TiO\textsubscript{2}(a), HZO(b), and CMO-HfO\textsubscript{2}(c) device stacks. 
    }
    \label{fig:article2-figs10}
\end{figure}

Even if analog switching could provide access to a quasi-continuous state distribution, each technology presents different switching dynamics that promote some discrete states. This effect is evidenced in \suppautoref{fig:article2-figs10}, where random pulses tend to promote extreme resistance states (HRS and LRS). Notably, the state's distribution in TiO\textsubscript{2} appears more continuous, while HZO and CMO/HfO\textsubscript{2} favor LRS and HRS, respectively. 

\section{BEOL integration}

\begin{figure}[H]
     \centering
         \includegraphics[clip,width=18cm,height=10cm,keepaspectratio, width=1\textwidth]{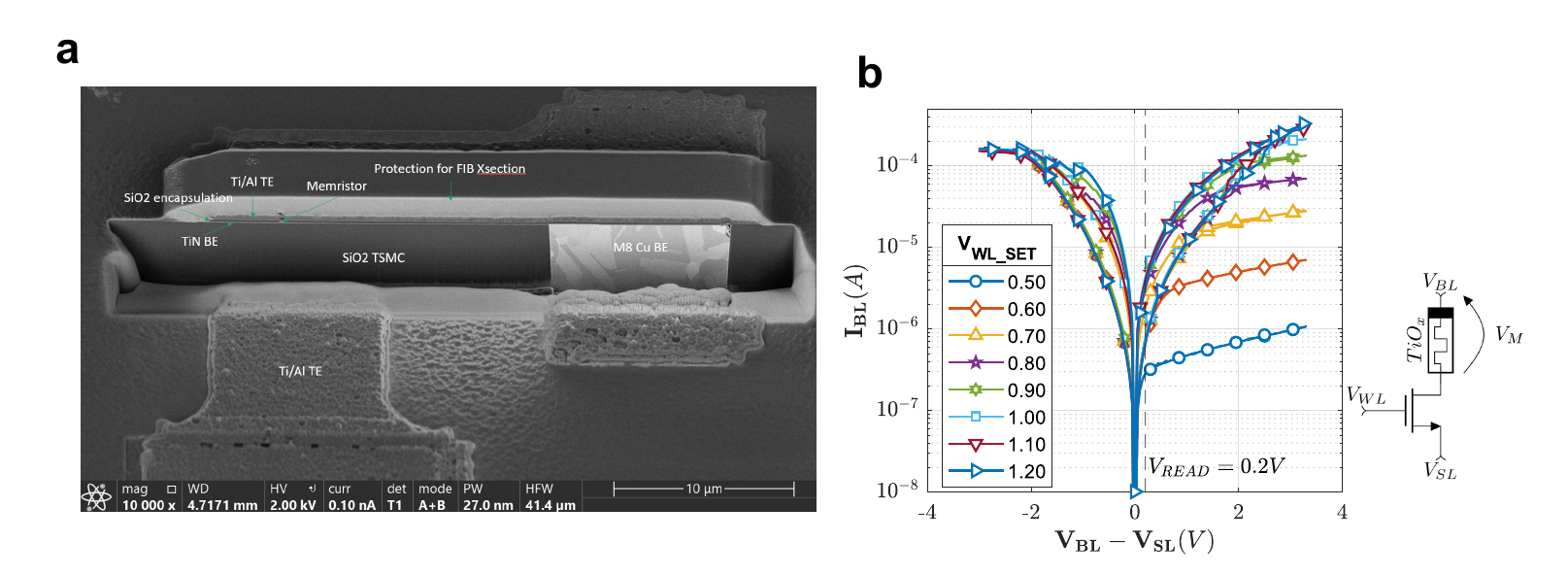}
    \caption{Back end of line (BEOL) integration of devices
    {\footnotesize: \textbf{a} Integration of TiO2-based memristors on CMOS chip for future demonstration of online learning depicted through a FIB cross-section. \textbf{b} Electrical characterization of the integrated device for different gate voltages (compliance current)}}
    \label{fig:article2-figs11}
\end{figure}

\end{document}